\documentclass[runningheads]{llncs}

\usepackage[mobile]{eccv}

\usepackage{eccvabbrv}

\usepackage{epsfig}
\usepackage{graphicx}
\usepackage{amsmath}
\usepackage{amssymb}
\usepackage{booktabs}
\usepackage{overpic}
\usepackage{multirow,tabularx}
\usepackage{gensymb}
\usepackage{pifont}
\usepackage{adjustbox}
\usepackage{array}
\usepackage{nicematrix}
\usepackage[usestackEOL]{stackengine}
\usepackage{xpatch}
\usepackage[font=small,skip=3pt,belowskip=-7pt]{caption}
\usepackage{placeins}

\definecolor{limegreen}{HTML}{badc58}
\definecolor{myyellow}{HTML}{f6e58d}

\renewcommand{\paragraph}[1]{\vspace{0.2mm}\noindent\textbf{#1}\:}
\newcommand{\cbest}[1]{\cellcolor{limegreen}\textbf{#1}}
\newcommand{\csecond}[1]{\cellcolor{myyellow}#1}

\renewcommand{\vec}[1]{\boldsymbol{#1}}
\newcommand{\mat}[1]{\mathbf{#1}}

\newcommand{\loss}{\mathcal{L}}

\newcommand{\image}{\mat{z}}
\newcommand{\denoisedimage}{\hat{\image}}
\newcommand{\cond}{\vec{y}}
\newcommand{\diffmodelweights}{\vec{\phi}}
\newcommand{\renderfnweights}{\vec{\theta}}
\newcommand{\kernelweights}{\vec{\psi}}
\newcommand{\eps}{\vec{\epsilon}}
\newcommand{\diffmodel}{\eps}

\newcommand{\timestep}{t}
\newcommand{\guidance}{\omega}

\newcommand{\bias}{b}
\newcommand{\approxbias}{\hat\bias}
\newcommand{\proj}{p}

\usepackage[pagebackref,breaklinks,colorlinks,citecolor=eccvblue]{hyperref}

\usepackage{orcidlink}

\begin{document}

\title{Score Distillation Sampling with \\ Learned Manifold Corrective} 

\titlerunning{LMC-SDS}

\author{Thiemo Alldieck\orcidlink{0000-0002-9107-4173} \and
Nikos Kolotouros\orcidlink{0000-0003-4885-4876} \and
Cristian Sminchisescu\orcidlink{0000-0001-5256-886X}}

\authorrunning{Alldieck et al.}

\institute{Google Research\footnote[1]{now at Google DeepMind}}

\maketitle

\begin{abstract}
\vspace{-4mm}
Score Distillation Sampling (SDS) is a recent but already widely popular method that relies on an image diffusion model to control optimization problems using text prompts.
In this paper, we conduct an in-depth analysis of the SDS loss function, identify an inherent problem with its formulation, and propose a surprisingly easy but effective fix.
Specifically, we decompose the loss into different factors and isolate the component responsible for noisy gradients.
In the original formulation, high text guidance is used to account for the noise, leading to unwanted side effects such as oversaturation or repeated detail.
Instead, we train a shallow network mimicking the timestep-dependent frequency bias of the image diffusion model in order to effectively factor it out.
We demonstrate the versatility and the effectiveness of our novel loss formulation through qualitative and quantitative experiments, including optimization-based image synthesis and editing, zero-shot image translation network training, and text-to-3D synthesis.
\vspace{-2mm}
\end{abstract}

\section{Introduction}
\vspace{-2mm}
\label{sec:intro}

Image diffusion models \cite{ho2020denoising} have recently become the \emph{de facto} standard for image generation.
Especially text-to-image models \cite{nichol2021glide,saharia2022photorealistic,rombach2022high, ramesh2022hierarchical} have  emerged as powerful tools for high fidelity, diverse, image synthesis.
By being controlled through natural language, these models are extremely easy to use and thus open up a wide range of creative applications, without the need for special training.
Beyond image synthesis, image diffusion models have been successfully deployed in applications like image restoration, inpatining, editing, super-resolution, or colorization \cite{chung2022diffusion,kawar2022denoising, meng2022sdedit,gao2023implicit}, among others.
Image diffusion models are typically obtained using very large training sets and thus inherently represent the distribution of natural images.
While applications typically make use of the generation process of diffusion models, \eg by inserting images into the process \cite{meng2022sdedit, lugmayr2022repaint} or by altering the denoising function \cite{chung2022improving, chung2022diffusion,kawar2022denoising}, relatively little research has been conducted on how diffusion models can be used as rich, general purpose image priors.
Score Distillation Sampling (SDS) proposed in DreamFusion \cite{poole2022dreamfusion} is one of the few exceptions.
The SDS loss formulation uses a pretrained text-to-image model \cite{saharia2022photorealistic} to measure how well a given text matches an observation.
In DreamFusion this loss is being used to generate 3D assets from textual descriptions, an idea that was quickly adopted \cite{wang2023score,lin2023magic3d,ruiz2023dreambooth,metzer2023latent,Liu_2023_ICCV,kolotouros2023dreamhuman} and essentially established a new research direction: text-to-3D synthesis.
While being mostly used in this context, the design of the SDS loss is by no means  only tailored to text-to-3D applications.
In fact, SDS is an image loss and can be used in much wider contexts  \cite{jain2023vectorfusion,hertz2023delta}.
However, in its original formulation the SDS loss may degrade the observation, be too eager to match the text prompt, or provide meaningless gradients which inject noise into the objective.
In this paper, we conduct an extensive analysis of the SDS loss, identify an inherent problem with its formulation, propose a surprisingly easy but effective fix, and demonstrate its effectiveness in several applications, including optimization-based image synthesis and editing, zero-shot training of image translation networks, and text-to-3D synthesis.
Concretely, our new loss formulation -- Score Distillation Sampling with Learned Manifold Corrective, or LMC-SDS for short -- aims to provide better gradients along the direction of the learned manifold of real images.
We present evidence that gradients towards the learned manifold are extremely noisy in SDS and that high text guidance is needed to compensate for the noisy signal. Further, we show that high guidance -- or lack thereof -- is a possible explanation for the artifacts observed with SDS, \cf \cref{sec:analysis}. %
In contrast, applications relying on our novel formulation benefit from meaningful manifold gradients, may use lower text guidance, and produce results of overall higher visual fidelity.

\begin{figure}[t]
    \centering
    \includegraphics[width=\linewidth]{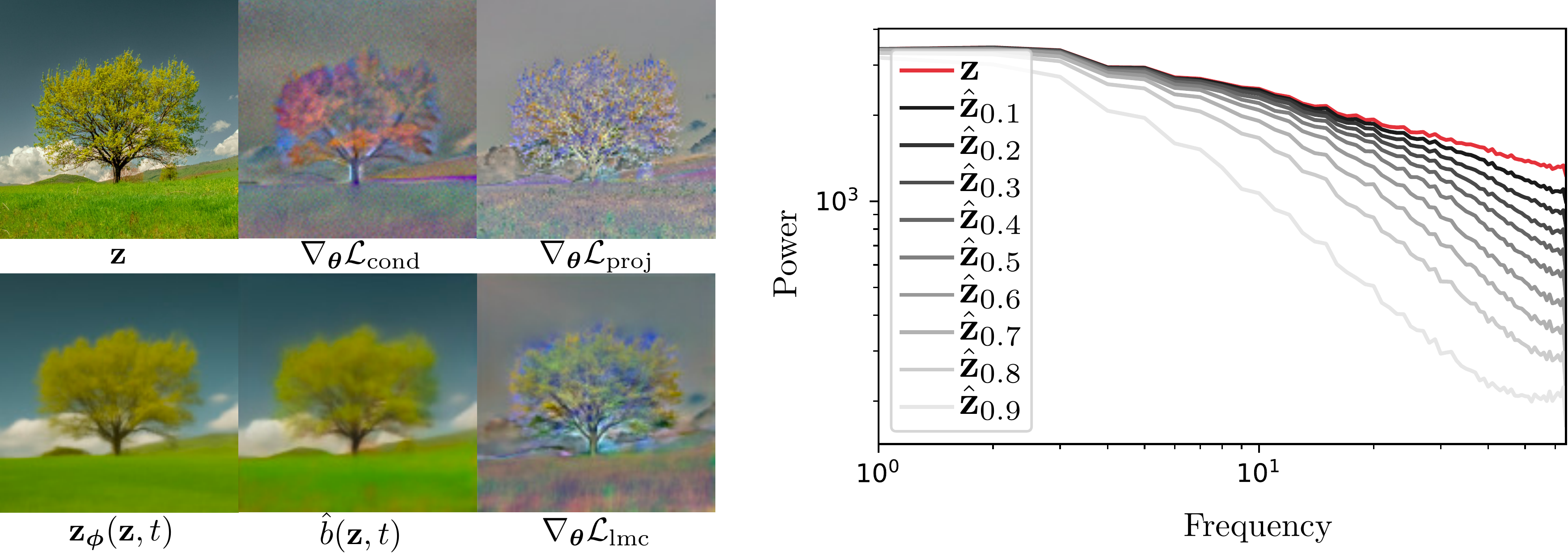}
    \caption{\textbf{Left:} Visualization of SDS and LMC-SDS gradients \wrt pixel values and estimated denoised images for the given image $\image$, the prompt $\cond=\text{\emph{``autumn''}}$, and $t=0.5$. We visualize the negative gradient, \ie the direction of change. \textbf{Right:} Power spectra of denoised images $\hat{\image}_\timestep$ for varying time-step $\timestep$ compared to the power spectrum of natural images $\image$. See \cref{sec:analysis} for details. }
    \label{fig:grad_vis}
    \vspace{-2mm}
\end{figure}

In summary, our contributions are: (1) conducting an in-depth analysis of the SDS loss, (2) identifying an inherent problem with the formulation, and (3) proposing a novel LMC-SDS loss, providing cleaner gradients and eliminating the need for high guidance. (4) In extensive experiments, we demonstrate the effectiveness and versatility of the new LMC-SDS loss.

\vspace{-3mm}
\section{Related Work}
\vspace{-3mm}

Methods that use image diffusion models as priors for tasks other than image synthesis can be broadly categorized into two lines of work.

The first class of methods leverage image diffusion networks to solve specific tasks by relying on the diffusion sampling process.
We discuss here a number of representative methods and refer the reader to \cite{li2023diffusion, croitoru2023diffusion} for detailed discussions.
DDRM \cite{kawar2022denoising} tackles linear inverse problems such as deblurring, super-resolution, colorization, or inpainting by defining a modified variational posterior distribution conditioned on the observation signal, which can then be used for sampling.
DPS \cite{chung2022diffusion} focuses on general inverse problems and proposes to augment the score prediction at each denoising step with an additional term that models the likelihood of the observations.
MCG \cite{chung2022improving} builds on top and improves the fidelity of generation by introducing additional manifold constraints. SDEdit \cite{meng2022sdedit} uses diffusion models for image editing by first adding noise to the input image and then denoising it using the text-to-image diffusion prior. RePaint \cite{lugmayr2022repaint} uses unconditional image diffusion models as a prior for solving image inpainting.
All these methods use pre-trained diffusion models. In contrast, methods like Palette \cite{saharia2022palette} or Imagic \cite{kawar2023imagic} require training a diffusion model from scratch or fine-tuning a pre-trained one, respectively.
This line of work exclusively deals with image editing problems and is typically restricted to the resolution of the diffusion models.

More related to our method is the second category:
This body of work uses pre-trained diffusion models within general purpose loss functions in iterative optimization settings.
Score Jacobian Chaining \cite{wang2023score} propagates the score of a text-to-image diffusion model through a differentiable renderer to supervise the generation of 3D models.
DreamFusion \cite{poole2022dreamfusion} takes an orthogonal approach and proposes the Score Distillation Sampling loss. SDS is a modification of the diffusion training objective that does not require expensive backpropagation through the diffusion model and encourages the alignment of the generated image with a conditioning signal.
DDS \cite{hertz2023delta} proposes an improved version of SDS, specifically designed for image editing and aims to reduce the artifacts introduced by noisy SDS gradients.
In similar spirit and in concurent work, NFSD \cite{katzir2023noisefree} aims to reduce noise in the gradients by adding negative conditioning.
In contrast, our method neither requires a \emph{source prompt} nor a \emph{negative prompt}.
SparseFusion \cite{zhou2023sparsefusion} introduces a multi-step SDS loss that performs multiple denoising steps instead of one, which adds significant computational overhead.
ProlificDreamer \cite{wang2023prolificdreamer} proposes a generalization of SDS from point estimates to distributions, but unlike our approach requires computationally expensive fine-tuning of a diffusion model during optimization.
Collaborative Score Distillation \cite{kim2023collaborative} generalizes SDS to multiple samples and is inspired by Stein variational gradient descent \cite{liu2016stein}.
Category Score Distillation Sampling (C-SDS) focuses on 3D generation and replaces the noise estimation error in SDS with a difference between the noise predictions of a standard and a multi-view consistent diffusion model \cite{zou2023sparse3d}. Unlike our method, C-SDS is only applicable to 3D generation tasks.
In contrast, we propose a general purpose loss function with very little computational overhead.

\vspace{-3mm}
\section{Analysis}
\vspace{-3mm}
\label{sec:analysis}

An image diffusion model is a generative model that is trained to reverse the diffusion process that transforms a natural image into pure noise \cite{ho2020denoising}.
An image is formed by iteratively removing small amounts of Gaussian noise over a fixed variance schedule $\bar\alpha_\timestep$ until an image is created.
Text-to-image models are additionally conditioned on text to steer the denoising process in a direction matching the textural description.
For a given image $\image$, its textual description $\cond$, a randomly sampled timestep $\timestep \sim \mathcal{U}(0, 1)$, and random Gaussian noise $\eps \sim \mathcal{N}(\vec{0}, \mat{I})$, a denoising model $\diffmodel_{\diffmodelweights}$ parameterized by $\diffmodelweights$ can be trained by minimizing the diffusion loss
\begin{equation}
    \loss_\text{diff} = w(t) \left\| \diffmodel^{\guidance}_{\diffmodelweights}(\image_\timestep, \cond, \timestep) - \eps \right\|^2_2,
\end{equation}
where $\image_\timestep = \sqrt{\bar\alpha_t}\image + \sqrt{1 - \bar\alpha_t} \eps$ refers to the noisy version of $\image$ and $w(t)$ is a weighting function, omitted in the sequel.
DreamFusion \cite{poole2022dreamfusion} showed that given a pre-trained model $\diffmodel_{\diffmodelweights}$ the diffusion loss can be utilized for optimization problems. 
For an arbitrary differentiable rendering function returning $\image$ and parameterized by $\renderfnweights$, the gradient of the denoising function \wrt $\renderfnweights$ is given by
\begin{equation}
    \nabla_{\renderfnweights} \loss_\text{diff} = \left ( \diffmodel^{\guidance}_{\diffmodelweights}(\image_\timestep, \cond, \timestep) - \eps \right )
    \frac{\partial \diffmodel^{\guidance}_{\diffmodelweights}(\image, \cond, \timestep)}{\partial \image_{\timestep}}
    \frac{\partial \image_{\timestep}}{\partial \renderfnweights}.
\end{equation}
In practice, the
Jacobian term is omitted to avoid backpropagating through the denoising model and the gradient is approximated as
\begin{equation}
\label{eq:grad_sds}
    \nabla_{\renderfnweights} \loss_\text{SDS} = \left ( \diffmodel^{\guidance}_{\diffmodelweights}(\image_\timestep, \cond, \timestep) - \eps \right )
    \frac{\partial \image_{\timestep}}{\partial \renderfnweights},
\end{equation}
resulting in the gradient of original SDS loss. We will now rewrite \cref{eq:grad_sds} and conduct an analysis of the derived components.

Using classifier-free guidance \cite{ho2021classifier}, the predicted noise is the sum of the $\cond$-conditioned and the unconditioned noise prediction weighted by the guidance weight $\guidance$
\begin{equation}
\label{eq:classifier_free_guidance}
\diffmodel^{\guidance}_{\diffmodelweights}(\image_\timestep, \cond, \timestep) = \guidance \diffmodel_{\diffmodelweights}(\image_\timestep, \cond, \timestep) + (1 - \guidance) \diffmodel_{\diffmodelweights}(\image_\timestep, \timestep),
\end{equation}
which we can rewrite as
\begin{equation}
\label{eq:classifier_free_guidance_rewrite}
    \diffmodel^{\guidance}_{\diffmodelweights}(\image_\timestep, \cond, \timestep) = \guidance \left(\diffmodel_{\diffmodelweights}(\image_\timestep, \cond, \timestep) - \eps_{\diffmodelweights}(\image_\timestep, \timestep)\right) + \diffmodel_{\diffmodelweights}(\image_\timestep, \timestep).
\end{equation}
By inserting \cref{eq:classifier_free_guidance_rewrite} into \cref{eq:grad_sds} we can express $\nabla_{\renderfnweights}\loss_\text{SDS}$ as
\begin{subequations}
\begin{equation}
\nabla_{\renderfnweights}\loss_\text{SDS} = \nabla_{\renderfnweights}\loss_\text{cond} + \nabla_{\renderfnweights}\loss_\text{proj}
\end{equation}
\begin{equation}
\nabla_{\renderfnweights}\loss_\text{cond} = \guidance \left(\eps_{\diffmodelweights}(\image_\timestep, \cond, \timestep) - \eps_{\diffmodelweights}(\image_\timestep, \timestep)\right) \frac{\partial \image_{\timestep}}{\partial \renderfnweights}
\end{equation}
\begin{equation}
\nabla_{\renderfnweights}\loss_\text{proj} = \left(\eps_{\diffmodelweights}(\image_\timestep, \timestep) - \eps \right)\frac{\partial \image_{\timestep}}{\partial \renderfnweights}.
\end{equation}
\end{subequations}
The two loss components can be interpreted as follows: $\loss_\text{cond}$ maximizes the agreement of the image with the text prompt by providing gradients towards images formed through conditioning on $\cond$. $\loss_\text{proj}$ performs a single denoising step and provides gradients informing how the image $\image$ (and thereby parameters $\renderfnweights$) should change such that the image can be better denoised. As the denoising function was trained to minimize the expected squared error between the denoised and the original data, $\nabla_{\renderfnweights}\loss_\text{proj}$ can be understood as the gradient direction towards the manifold of natural images.

DreamFusion \cite{poole2022dreamfusion} originally proposed to set $\guidance=100$. This means that, following our derivation, $\loss_\text{cond}$ is weighted proportionally much higher than $\loss_\text{proj}$. This can be problematic as large guidance weights have been identified to be an important cause of over-exposure in diffusion models \cite{lin2023common}. In fact, DreamFusion produces detailed results but also has a tendency towards unrealistically saturated colors, see also \cref{fig:guidance_ablation}. On the other hand, setting $\guidance$ to lower values has been reported to result in very blurry images \cite{hertz2023delta}, a behavior that we also demonstrate in \cref{sec:experiments}.
Both behaviors, (1) the tendency to produce overly saturated results when using high guidance as well as (2)  blurry results for low guidance, can be explained by looking at the components of $\loss_\text{SDS}$ in isolation.
To this end, we consider the case where the rendering function is a no-op and simply returns pixel values, \ie $\image = \renderfnweights$. In \cref{fig:grad_vis} (left), we visualize $\nabla_{\renderfnweights}\loss_\text{cond}$ and $\nabla_{\renderfnweights}\loss_\text{proj}$ for this case. $\nabla_{\renderfnweights}\loss_\text{cond}$ returns the direction towards an image with altered colors. However, as $\nabla_{\renderfnweights}\loss_\text{cond}$ is not anchored in the manifold of natural images, the gradient may eventually point away from that manifold. $\nabla_{\renderfnweights}\loss_\text{cond}$ will always point towards changes that ``better'' respect the prompt  and eventually produce saturated colors or other artifacts (\cf \cref{sec:experiments}).
$\nabla_{\renderfnweights}\loss_\text{proj}$, on the other hand, incorrectly marks high-frequency detail originating from $\image$ for removal -- the reason for blurriness. 
This behavior can be understood when rewriting $\nabla_{\renderfnweights}\loss_\text{proj}$ as 
\begin{equation}
\label{eq:loss_proj_x0}
\nabla_{\renderfnweights}\loss_\text{proj} = \left(\frac{\sqrt{\alpha_t}}{\sqrt{1-\alpha_t}} \left(\image -\image_{\diffmodelweights}(\image_\timestep, \timestep)\right)\right)\frac{\partial \image_{\timestep}}{\partial \renderfnweights},
\end{equation} 
which is equivalent to comparing the predicted and true noise. By looking at \cref{eq:loss_proj_x0}, we can now see a possible explanation why $\loss_\text{proj}$ might not be optimal: $\denoisedimage_\timestep = \image_{\diffmodelweights}(\image_\timestep, \timestep)$ is an approximation of $\image$ and most probably not a perfect replica of $\image$, even when $\image$ already lies on the natural image manifold, see \cref{fig:grad_vis} (left). This is especially the case for large values of $\timestep$, where the denoising model has to reconstruct the true image from almost pure noise. More formally, we can observe a frequency bias in $\image_{\diffmodelweights}$ dependent on $\timestep$ when comparing images $\image$ with their denoised counterparts $\denoisedimage_\timestep$, see \cref{fig:grad_vis} (right).

We conclude the following: (1) The reason for oversaturated colors and other artifacts is due to $\loss_\text{cond}$ being ``over-eager'' to maximize the agreement of the image with the conditioning variable. (2) $\loss_\text{proj}$ provides deficient gradients ultimately removing high-frequency detail instead of ``correcting'' the gradient towards on-manifold solutions due to a frequency bias in $\image_{\diffmodelweights}$.
Thus, $\loss_\text{proj}$ appears to be a non-optimal choice and we provide an alternative  in the sequel. But first we use our decomposition to discuss other approaches to ``fix'' the SDS loss.

DDS \cite{hertz2023delta} proposes to use the difference between two SDS losses for image editing, one computed on the current state \wrt the target prompt and one computed on the initial image \wrt a textual description of that image. 
By computing an SDS difference, DDS entirely removes $\loss_\text{proj}$ and introduces a negative gradient pointing away from the initial state.
However, the need for an initial image is limiting possible use-cases.
In concurrent work, NFSD \cite{katzir2023noisefree} aims to obtain better results by replacing $\loss_{\text{proj}}$ with a $\loss_{\text{cond}}$ variant using \emph{negative} prompts for most timesteps, and a loss based on $\eps$ for small $\timestep$.
While the root for blurry results is removed by eliminating $\loss_\text{proj}$, both DDS and NFSD do not anchor the optimization along the learned manifold. 
Other methods aim for a better noise prediction, where by improving $\eps_{\diffmodelweights}(\image, \timestep)$, one may also improve $\loss_\text{proj}$ ($\loss_\text{proj}$ is the difference between the predicted and true noise).
\Eg SparseFusion \cite{zhou2023sparsefusion} introduces multi-step SDS. Hereby, instead of denoising in one step, up to 50 steps are used. While this may produce more informative gradients in $\loss_{\text{proj}}$, this is also significantly more expensive.
Through different means but towards the same goal, the concurrent work  ProlificDreamer \cite{wang2023prolificdreamer} learns a low-rank adaptation of the text-to-image model.
While this approach is effective, it also adds the overhead of fine-tuning an additional diffusion model.
Finally, some recent methods aim to alleviate artifacts by augmenting the original SDS formulation with advanced schedules for sampling $\timestep$ or with negative prompts \cite{armandpour2023re,shi2023MVDream}.
In contrast, we aim for a generally applicable solution.

\vspace{-3mm}
\section{Method}
\vspace{-3mm}
\label{sec:method}
We now present our solution to eliminating the frequency bias in $\loss_\text{proj}$ that we identified earlier.
In an attempt to better understand the behavior of the denoising model given $\image$ and $\timestep$, we model it as a two step process, as follows
\begin{equation}
\image_{\diffmodelweights}(\image_\timestep, \timestep) = \bias \circ \proj(\image_\timestep, \timestep) = \bias(\tilde\image, \timestep),
\end{equation}
where $\proj$ projects $\image$ onto the learned natural image manifold, $\tilde\image$ is the projected image, and $\tilde\image = \image$ for images on the manifold. $\bias$ is the error or frequency bias of the denoising model introduced by the current time step $\timestep$. 
In this two-step model we are only interested in $\proj$ and would like to neglect any gradients coming from $\bias$.
As it is unfeasible to correct for $\bias$, we propose to cancel out effects of $\bias$ by approximating $\bias$ with $\approxbias$ and applying $\approxbias$ on $\image$ before comparing $\image$ with $\denoisedimage_\timestep$:
\begin{equation}
\label{eq:lmc}
\small
\nabla_{\renderfnweights}\loss_\text{lmc} = \left(\approxbias(\image, \timestep) -\denoisedimage_\timestep\right) \frac{\partial \image_{\timestep}}{\partial \renderfnweights}
\end{equation}
In this formulation we omit both the Jacobian term from the denoising model (like in the original formulation, \cf \cref{eq:grad_sds}) as well as the Jacobian from $\approxbias$. Also, we empirically found that by dropping the weighting term  $\frac{\sqrt{\alpha_t}}{\sqrt{1-\alpha_t}}$ we can further improve visual fidelity -- which is similar to the common practice of predicting the original image instead of the noise when learning the denoising model. As demonstrated in \cref{fig:grad_vis} (left), given an adequate model $\approxbias$, $\nabla_{\renderfnweights}\loss_\text{lmc}$ focuses on global changes, whereas $\nabla_{\renderfnweights}\loss_\text{proj}$ is dominated by high-frequent detail. We now explain how to obtain $\approxbias$.

We propose to model $\approxbias$ as a learnable function $\approxbias_{\kernelweights}(\image, \timestep)$ parameterized by  $\kernelweights$.
We can observe $\bias$ as a function of $\timestep$ by sampling triplets $(\image, \denoisedimage_\timestep, \timestep)$, where $\image$ are random natural images assumed to be approximately on manifold images, \ie $\image\approx\tilde\image$.
Using these triplets we can learn $\approxbias_{\kernelweights}$ by minimizing $\| \denoisedimage_\timestep - \approxbias_{\image, \timestep} \|^2_2$, with $\approxbias_{\image, \timestep} = \approxbias_{\kernelweights}(\image, \timestep)$.
Modeled with a deterministic model, $\approxbias_{\kernelweights}$ is expected to under-fit the highly complex and probabilistic nature of $\bias$.
In fact, for a given pair $(\image, \timestep)$ at best it can learn the mean over all possible $\denoisedimage_\timestep$.
This is desired, because even after approximating $\bias(\image, \timestep)$ through $\approxbias_{\kernelweights}(\image, \timestep)$, we will be able to observe a gradient in $\loss_\text{lmc}$: a gradient towards a specific instance of $\denoisedimage_\timestep$.
Further, $\approxbias_{\kernelweights}$ only learns to ``blur'' images and neither generates new content nor is informed by the diffusion models' image manifold.
Thus, it will perform similar operations for on-manifold and off-manifold images and  $\loss_\text{lmc}$ may correct non-manifold images.
However, naively applying a $\approxbias_{\kernelweights}$ in \cref{eq:lmc}, learned that way, is causing an undesired effect, namely a drift in image statistics. 
Concretely we observed that images produced by $\approxbias_{\kernelweights}$ tend to underestimate the true image dynamics, especially for large $\timestep$ where $\denoisedimage_\timestep$ may be vastly different to $\image$.
By underestimating image dynamics, however, $\nabla_{\renderfnweights}\loss_\text{lmc}$ contains a gradient to ``correct''
for that and we have found images obtained via optimization using this gradient to be overly saturated and contrasting.
To this end, we make $\approxbias_{\kernelweights}$ agnostic to the global image statistics by computing the training loss in a normalized space
\begin{equation}
    \small
    \loss_\text{k} = \left\| \denoisedimage_\timestep - \left(\frac{\sigma({\denoisedimage_\timestep})}{\sigma(\approxbias_{\image, \timestep})} \left(\approxbias_{\image, \timestep} - \mu(\approxbias_{\image, \timestep})\right) + \mu(\approxbias_{\image, \timestep})\right) \right\|^2_2,
\end{equation}
with standard deviation $\sigma$ and mean $\mu$.
Predicting in a normalized space naturally requires to perform the same (normalisation) operation  in \cref{eq:lmc}, which we assume to be absorbed into $\approxbias$ as follows. %

\paragraph{Increasing Sample Diversity.}
The original SDS formulation has been reported to have mode-seeking behaviour. Consequently, optimization results obtained via $\loss_{\text{SDS}}$ have the tendency to look very similar, even when based on different random seeds.
We hypothesize this is due to optimizing over many steps and hence averaging the effects of varying noise $\eps$. To prevent such averaging behaviour, we propose to fix $\eps$ over the course of optimization -- which is in spirit similar to DDIM sampling \cite{song2020denoising}. Empirically, we found that optimizing with fixed $\eps$ is often unstable in the original SDS formulation.
We hypothesize this is rooted in $\loss_{\text{proj}}$ not providing proper gradients towards the image manifold and at the same time $\loss_{\text{cond}}$ providing too strong gradients towards prompt agreement. 
The optimization state may leave the manifold of natural images and drift towards a state associated more to an ``adversarial example'' for the given prompt.
While our loss formulation allows the use of a fixed $\eps$, probably due to both the usage of smaller $\guidance$ and better manifold correctives, we observe that we can further robustify the formulation by only fixing $\eps$ in $\loss_{\text{cond}}$, while continue to sample $\eps$ in $\loss_{\text{lmc}}$. This results in high quality, yet diverse, results, \cf \cref{sec:experiments}.

\paragraph{Implementation Details.} 
We use an image diffusion model producing images of $128 \times 128$ pixel resolution and trained on internal data sources \cite{saharia2022photorealistic} as the pre-trained text-to-image diffusion model $\diffmodel^{\guidance}_{\diffmodelweights}$ or $\image^{\guidance}_{\diffmodelweights}$, respecively.
We model $\approxbias_{\kernelweights}$ as a standard U-Net with four encoder layers with two Conv/ReLU/MaxPool blocks per layer, and skip connections to the decoder.
The first layer contains 32 filters and the number of filters is doubled in each layer in the encoder and halved in the decoder.
$\approxbias_{\kernelweights}$ has been trained until convergence, using $\loss_{\text{k}}$ as the only loss. For creating training triplets $(\image, \denoisedimage_\timestep, \timestep)$, we sampled random images $\image$ from OpenImages \cite{OpenImages}, random time-steps $\timestep \sim \mathcal{U}(0.02, 0.98)$, and passed both into $\image_{\diffmodelweights}$ obtaining $\denoisedimage_\timestep$.
We train $\approxbias_{\kernelweights}$ once and keep it fixed for all experiments.

\section{Experiments}
\vspace{-2mm}
\label{sec:experiments}

\begin{figure}[t]

    \centering
    \includegraphics[width=\linewidth]{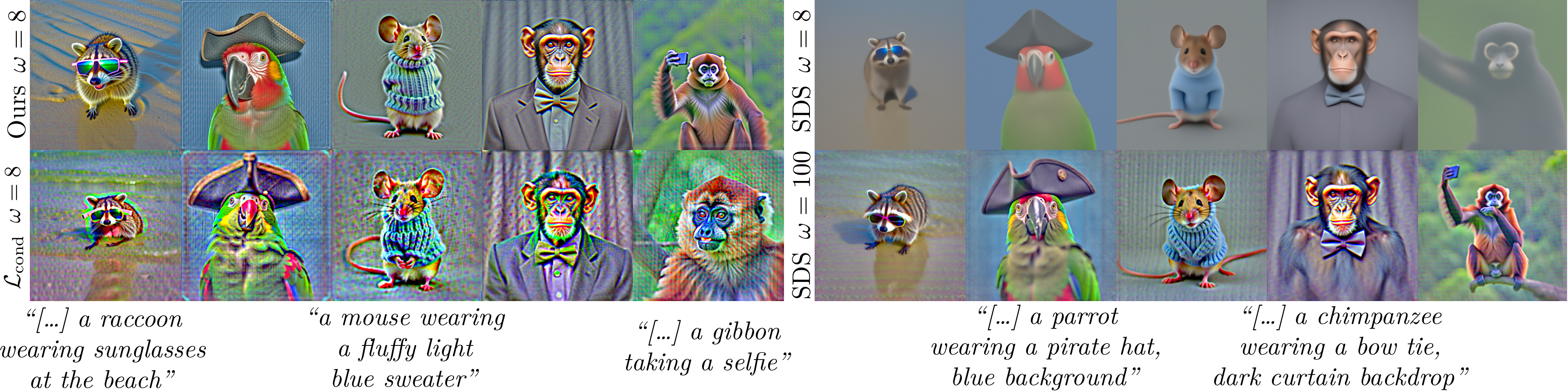}
    \caption{Optimization-based image synthesis results. We optimize an empty image to match a given prompt using our LMC-SDS, the original SDS loss, and $\loss_{\text{cond}}$. SDS struggles to create detailed content when using low guidance $\guidance$. High $\guidance$ produces detailed results but colors may be oversaturated (chimpanzee face), fake detail may appear (2nd mouse tail), or artifacts emerge. $\loss_{\text{cond}}$ is unstable to optimize and produces unrealistic colors. In contrast, our method produces detailed results with balanced colors.}
    \label{fig:compare_image_gen}

\end{figure}

We evaluate our proposed LMC-SDS loss in a number of qualitative and quantitative experiments, including image synthesis, image editing, image-to-image translation network training, and 3D asset generation.

\paragraph{Set-up.} If not noted otherwise, we scale the total loss by $\frac{1}{\guidance}$ to be able to ablate $\guidance$ without changing the gradient magnitude.
Fixed $\eps$ sampling is only used when explicitly stated.
The pre-trained diffusion model we use in our experiments produces images of $128\times128$px resolution.
We enable arbitrary higher resolutions in our image experiments through the following strategy: 
We downsample the optimization state to $128\times128$px for the first iterations.
This ensures that the whole image is affected by the loss, but the signal may be blurry.
Accounting for this, we sample random patches of arbitrary resolutions, from the image, for the remaining optimization steps.
To prevent the loss from producing fake detail in sampled image patches, we lower the learning rate during this phase through Cosine decay.

\paragraph{Baselines.}
This paper presents a generic novel loss function which uses an image diffusion model.
To this end, we compare with the original SDS formulation and with related work focused on improving SDS.
Specifically, we compare against DDS \cite{hertz2023delta}, NFSD \cite{katzir2023noisefree}, and multi-step SDS (MS-SDS) \cite{zhou2023sparsefusion}. %
DDS, NFSD, and MS-SDS originally were proposed using a latent diffusion model \cite{rombach2022high} and operate in latent space.
For fair comparisons, we adapted all methods to operate in pixel space and to use the same image diffusion model.
Note that optimizing in the (ambient) pixel space is generally considered harder due to higher-dimensionality.
Finally, we evaluate the effectiveness of our LMC-SDS loss for 3D asset generation using DreamFusion.
Hereby we focus on the improvements of our loss and deliberately exclude recent improvements over DreamFusion rooted in other factors, \eg novel-view synthesis networks or additional perceptual losses.
While being important findings, those factors are orthogonal and complementary to our approach.

\paragraph{Metrics.} We use the following metrics for evaluating our methods and baselines: We report Learned Perceptual Image Patch Similarity (LPIPS $\downarrow$) \cite{zhang2018perceptual} \wrt the source image, thus capturing how much an image was altered. Note that high LPIPS can be a result of either successful editing or image corruption.
To disambiguate these, we additionally report the CLIP score ($\uparrow$) \cite{hessel2021clipscore} \wrt the target prompt.
As CLIP can be insensitive, we additionally report CLIP-R-Precision ($\uparrow$) \cite{park2021benchmark}, the retrieval accuracy using the CLIP model, where feasible.

\begin{figure}[t]
    \centering
    \includegraphics[width=\linewidth]{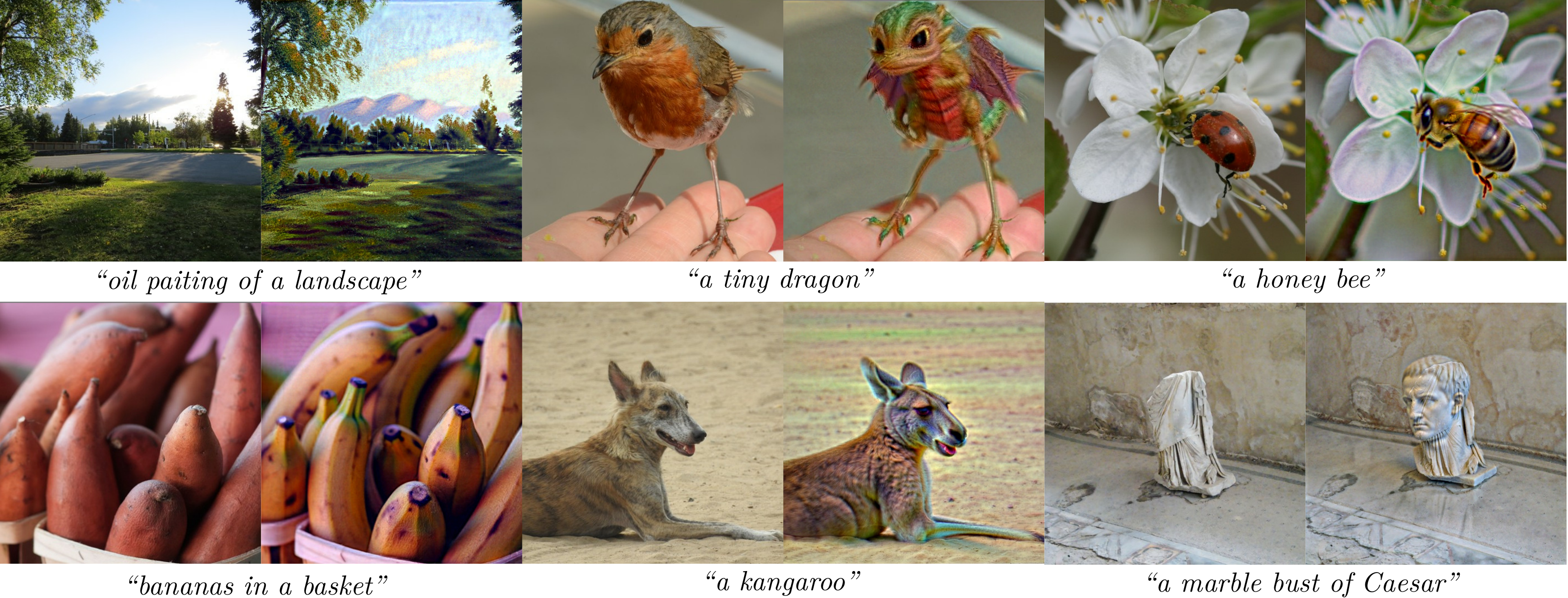}
    \caption{Examples of optimization-based image editing results. We show pairs of input images (left) and editing result (right).}
    \label{fig:edit}
\end{figure}

\begin{figure}[t]
    \centering
    \includegraphics[width=\linewidth]{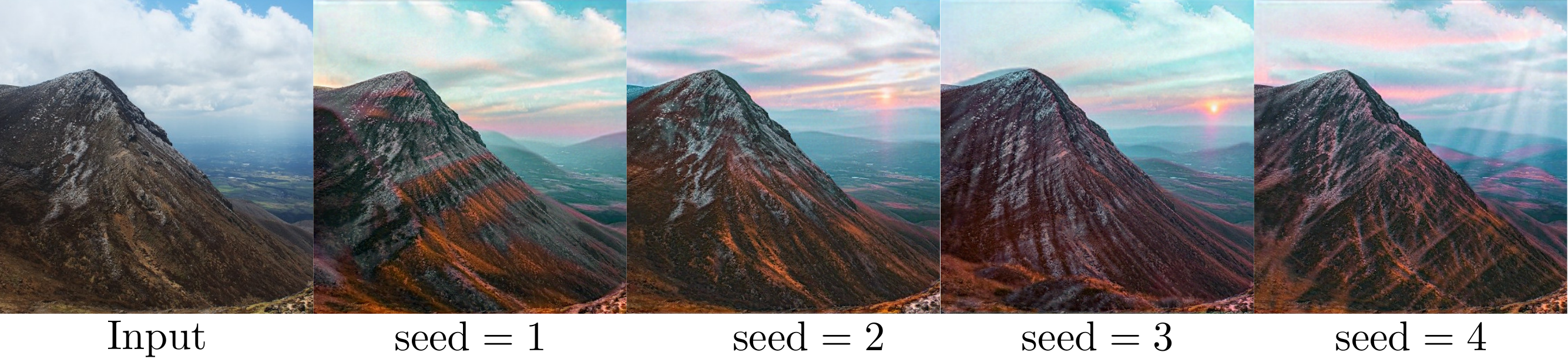}
    \caption{By fixing $\eps$ in $\loss_{\text{cond}}$ we can obtain diverse editing results. We show four variants of optimization-based editing results of the input image under the given prompt \emph{``a mountain during sunset''}.}
    \label{fig:edit_seed}
\vspace{-2mm}
\end{figure}

\subsection{Image Synthesis \& Editing}
\vspace{-2mm}

We first explore our LMC-SDS loss for optimization-based image synthesis and image editing.

\paragraph{Synthesis.} 
To be able to evaluate and ablate the original SDS loss and our LMC-SDS in isolation (without additional interfering losses), we propose the task of optimization-based image synthesis.
Instead of using the denoising schedule of a diffusion model, we initialize an image with 50\% grey and \emph{optimize} its pixel values to match a given prompt, following the patching strategy described earlier.
While it is not expected that images obtained using this strategy are on par with those obtained via the generation process, the results allow to draw meaningful conclusions about the properties of the loss functions.
We show results of our loss and SDS in \cref{fig:compare_image_gen}.
We additionally include results obtained by only using $\loss_\text{cond}$, ablating the influence of our proposed $\loss_\text{lmc}$. 
Optimizing solely with $\loss_\text{cond}$ is unstable and the final images feature clearly visible defects, such as oversaturated colors or noisy patterns.
The prompt is also not always well respected as \eg in the case of the gibbon.
The original SDS formulation produces blurry and desaturated images when using low guidance $\guidance$.
High $\guidance$ leads to better performance but artifacts are present, \eg fake detail like the second mouse tail, oversaturated colors like in the chimpanzee's face, or blocky artifacts.
In contrast, our method produces images of higher visual fidelity which respect the prompt well.
Additionally, we empirically found that problems of the individual losses become more pronounced when optimizing longer (here we only optimize for 500 steps, \cf \cref{sec:zeroshot} for effects after 50K steps).

\paragraph{Editing.}
Our loss can also be used for optimization-based image editing.
We load existing natural images and optimize their pixel values directly.
Optionally, we may constrain the optimization with an L2 loss \wrt the original pixel values.
In \cref{fig:edit} (more in  Supp.\ Mat.) we show qualitative results of so edited images along with the corresponding target prompt.
All results faithfully respect the target prompt while keeping the image structure intact.
We can also vary the result by fixing $\eps$ over the course of the optimization, \cf \cref{sec:method}.
In \cref{fig:edit_seed}, we show an input image along with four result variants.
All results respect the target prompt well, while at the same time being significantly different from each other.

\begin{figure}[t]
    \centering
    \resizebox{\linewidth}{!}{%
    \includegraphics[width=0.428\linewidth]{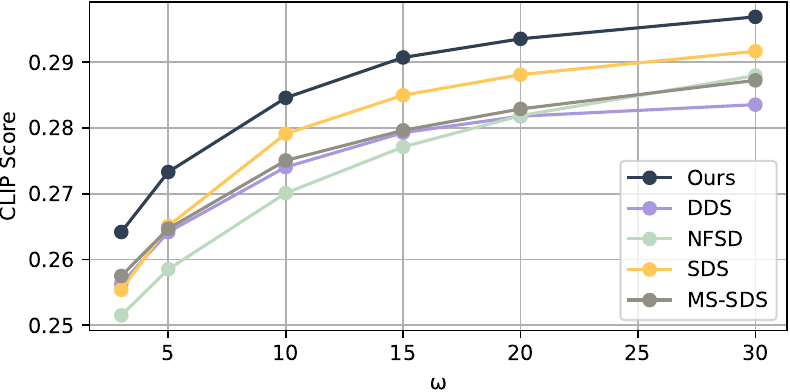}
    \hspace{2mm}
    \includegraphics[width=0.571\linewidth]{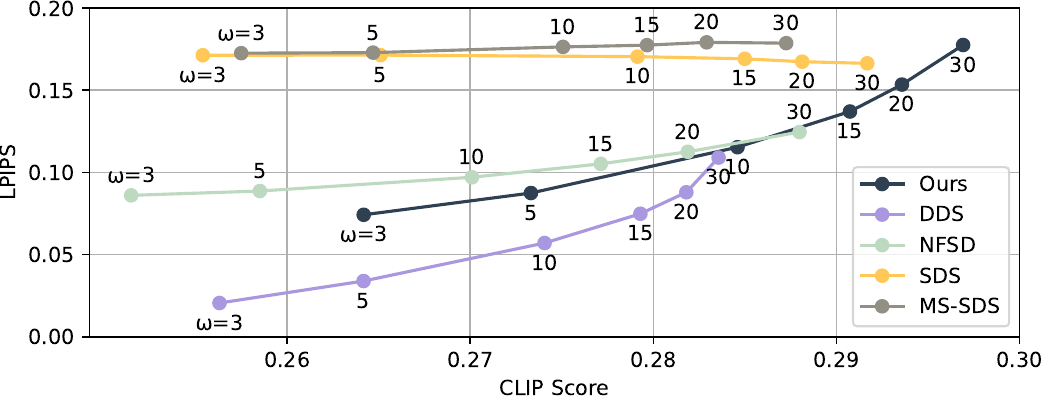}
    }
    \caption{Quantitative results for optimization-based editing under varying $\guidance$. \textbf{Left}: Our method results in the highest CLIP score over all baselines for all $\guidance$. \textbf{Right:} We plot LPIPS over CLIP for further performance insights: DDS stays close to the original image (lowest LPIPS) by performing only small edits (low CLIP). SDS and MS-SDS respect the prompt better (higher CLIP), but corrupt the image (high LPIPS). NFSD corrupts the image less (lower LPIPS), but exhibits weak editing capabilities (low CLIP). Our method shows the strongest editing capabilities (highest CLIP), while staying close to the original structure.}
    \label{fig:edit_plot}

 \vspace{4mm}
    \centering
    \includegraphics[width=0.75\linewidth]{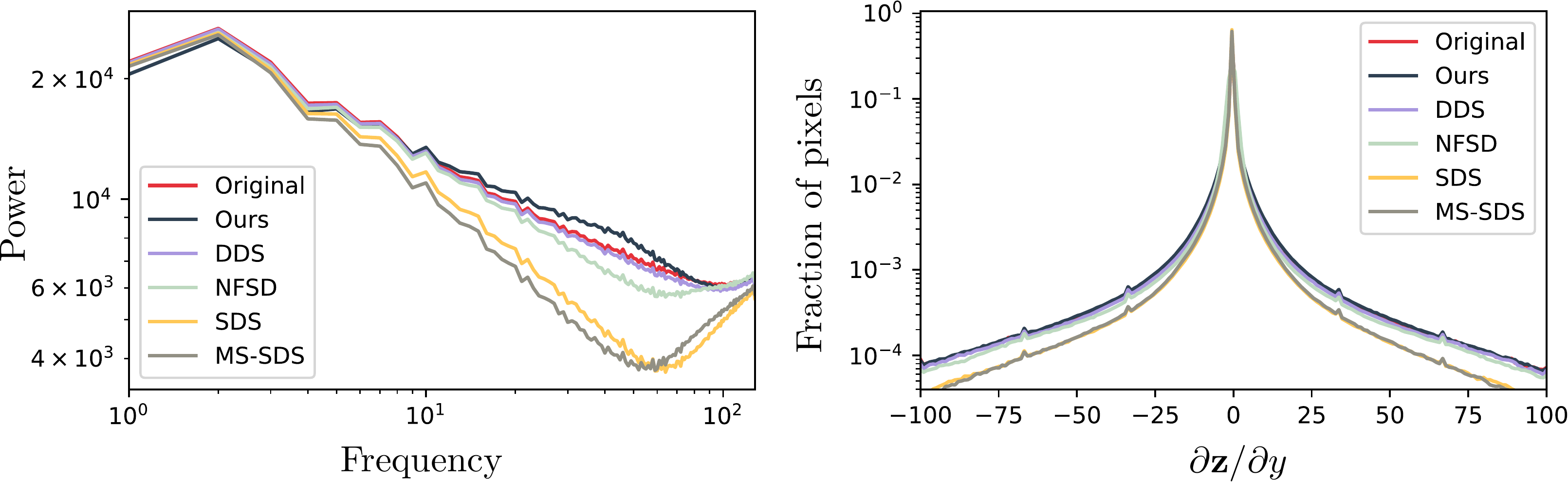}
    \caption{Comparison of image statistics before (Original) and after editing with $\guidance=5$. We plot the average power spectra (left) and vertical derivative histograms  (right). Ours, DDS, and NFSD preserve the image statistics well, while SDS-based methods introduce significant blur.}
    \label{fig:statistics}
    \vspace{-2mm}
\end{figure}

\begin{figure}[t]
    \centering
    \includegraphics[width=\linewidth]{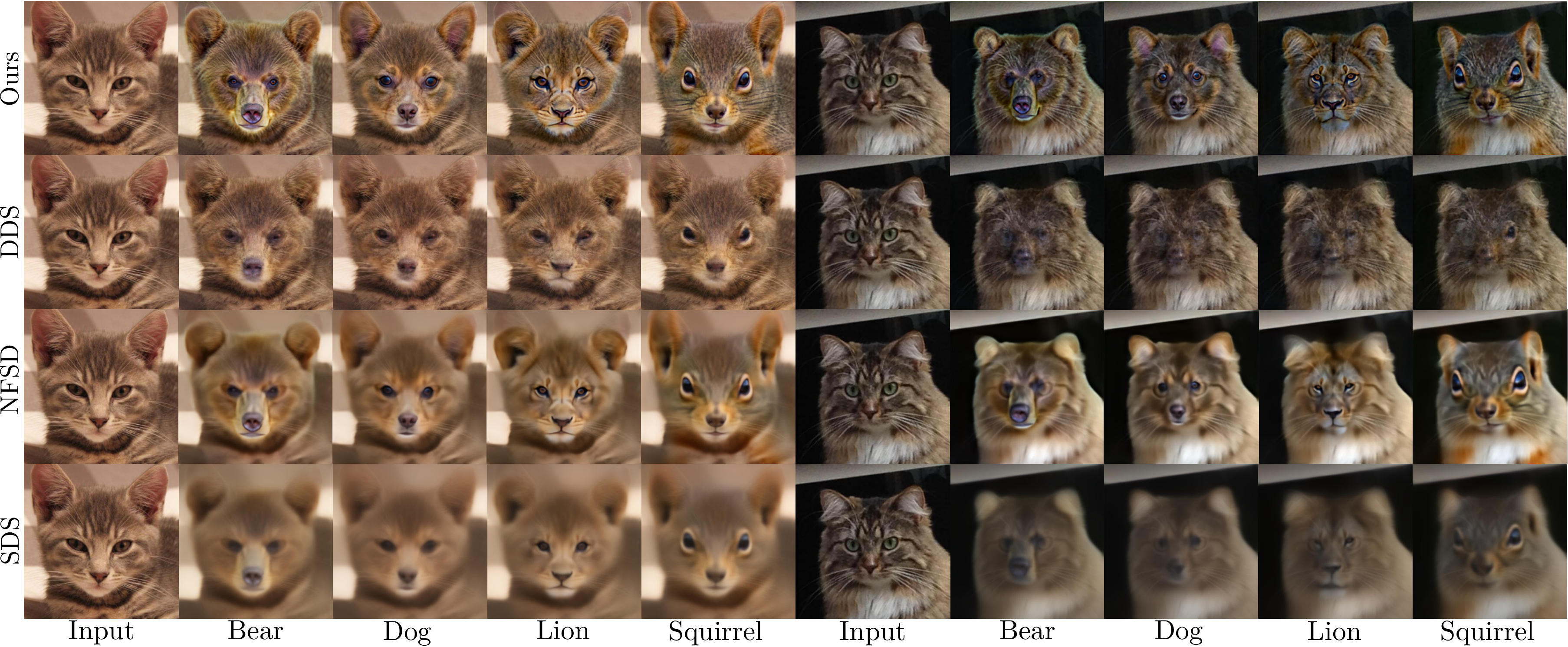}
    \caption{Results obtained with \emph{cats-to-others} image translation networks trained with different losses. The network trained with SDS and NFSD fail to create detailed results. The DDS network produces rather uniform results, whereas the network trained with our LMC-SDS creates realistic and detailed images.}
    \label{fig:compare_zero_shot}

\end{figure}

\begin{table}[t]

    \centering
    \begin{tabular}{l|ccc}
         & LPIPS $\downarrow$ & CLIP Score $\uparrow$ &  CLIP-R-Precision $\uparrow$ \\ \hline
    DDS  & \cbest{0.143}  & 0.253 & 0.333  \\ 
    NFSD  &  0.332 & 0.255 &  0.434 \\ 
    SDS & 0.401 & 0.256 & 0.355  \\ 
    SDS, $\guidance=100$ &  0.201 & \csecond{0.259}  &  \csecond{0.482}  \\ \hline
    Ours & \csecond{0.178}  & \cbest{0.261} & \cbest{0.592}           
    \end{tabular}
    \vspace{1mm}
    \caption{Numerical results of the \emph{cats-to-others} network training experiment. Note that LPIPS is computed \wrt the source image, meaning a low score may mean the image has not been edited enough. This is the case for DDS, as evident from low R-Precision.}
    \label{tab:numbers_zero_shot}
    \vspace{-10mm}
\end{table}

For a numerical evaluation, we use an internal dataset of 1,000 images with text editing instructions in form of source and target prompts, similar to the dataset used in InstructPix2Pix \cite{brooks2022instructpix2pix}.
We optimize images to match the target prompt solely using our LMC-SDS and baselines DDS, NFSD, SDS, and MS-SDS.
For MS-SDS we chose to use five step denoising.
We also experimented with more steps, but didn't observe significantly different performance.
We report numerical results with varying $\guidance$ (without loss rescaling) in \cref{fig:edit_plot}.
We found DDS to be least effective and producing results closest to the original image.
We hypothesize that the negative gradient \wrt to the original image may not always be helpful.
This could also explain a behavior reported by original authors, where editing results on real images are ``inferior'' over those on synthetic images. A possible reason being, that since for synthetic images the original prompts are describing the image well, the negative gradients are more helpful in that case.
SDS can effectively edit the image, but also corrupts the image. 
This is also evident from the altered image statistics, see \cref{fig:statistics}. NFSD corrupts the image less, but exhibits weak editing capabilities. 
In our experiments, MS-SDS performs very similar to SDS, corrupting the image in similar manner, but also being slightly less effective for editing.
MS-SDS produces clear and detailed denoised images as part of the loss computation.
We hypothesize that since denoised images and the current optimization state may differ drastically, and as denoised images vary considerably at each step, the computed gradients are not always very meaningful.
The original paper proposes MS-SDS to be used with a view-conditioned diffusion model, where the denoised images probably vary much less and computed gradients may be more directed.
Our loss respects the prompt best (highest CLIP), and finds a good balance between preserving the original structure and performing significant edits.

To quantify the how much the result diversity can be increased through fixing $\eps$, we re-ran the experiment using LMC-SDS ($\guidance = 30$) ten times with different seeds, both using the standard strategy and with fixing $\eps$ in $\loss_\text{cond}$. We then computed the LPIPS (lower is more similar) of nine runs \wrt to the 10th (reference) run.
The results using the standard strategy are all very similar with a per-run average of $0.041 \pm 0.007$ \wrt the reference run. In contrast, our new strategy produces much more diverse results: $0.158 \pm 0.018$.

\vspace{-2mm}
\subsection{Image-to-image Translation Network Training}
\label{sec:zeroshot}
\vspace{-2mm}
Optimization-based editing is not always an optimal choice, \eg when building interactive editing tools where execution speed matters.
Further, optimization can be unstable or perform better or worse, depending on the input images or target prompts.
To this end, we explore to what extend LMC-SDS can be used to train a reliable and fast image-to-image translation network.
Given a known distribution of images, we train a network to translate images from that distribution to an unseen target distribution, only described by a target caption.
Inspired by the authors of DDS \cite{hertz2023delta}, we train a \emph{cats-to-others} network, with ``others'' being bears, dogs, lions, and squirrels.
The network transforms an image of a cat into an image of one of the other species.
In DDS, the authors fine-tune a pre-trained diffusion model into a single-step feedforward model using images created by that same diffusion model.
In contrast, we train a standard U-Net from scratch and use a dataset of real images, a much more practical and versatile approach.
Concretely, we train using cat images from AFHQ \cite{choi2020starganv2} and evaluate on 5\% held out test images.
We use the target prompt \emph{``a photo of a [class]''} with $\guidance=15$ for all methods and the source prompt \emph{``a photo of a cat''} for DDS.
Moreover, we stabilize the training using an L2 loss on the original image and use Cosine scheduling to let the network first focus on reconstructing the original image and then on edits, similarly to the experiment in DDS.
The class is passed via a frequency encoded class id concatenated to each pixel. 
We report numerical results in \cref{tab:numbers_zero_shot} and show qualitative examples in \cref{fig:compare_zero_shot}. Additional analysis is provided in the Suppl.\ Mat.
SDS performs similarly to previous experiments for this task.
While the target class is evident in the resulting images, the results feature strong deficiencies.
Again DDS struggles with significant edits and stays close to the source.
In contrast, the network trained with our LMC-SDS loss produces realistic and convincing edits.
This is also evident from the significantly higher CLIP-R-Precision among all network variants.

\begin{figure}
    \centering
    \includegraphics[width=\linewidth]{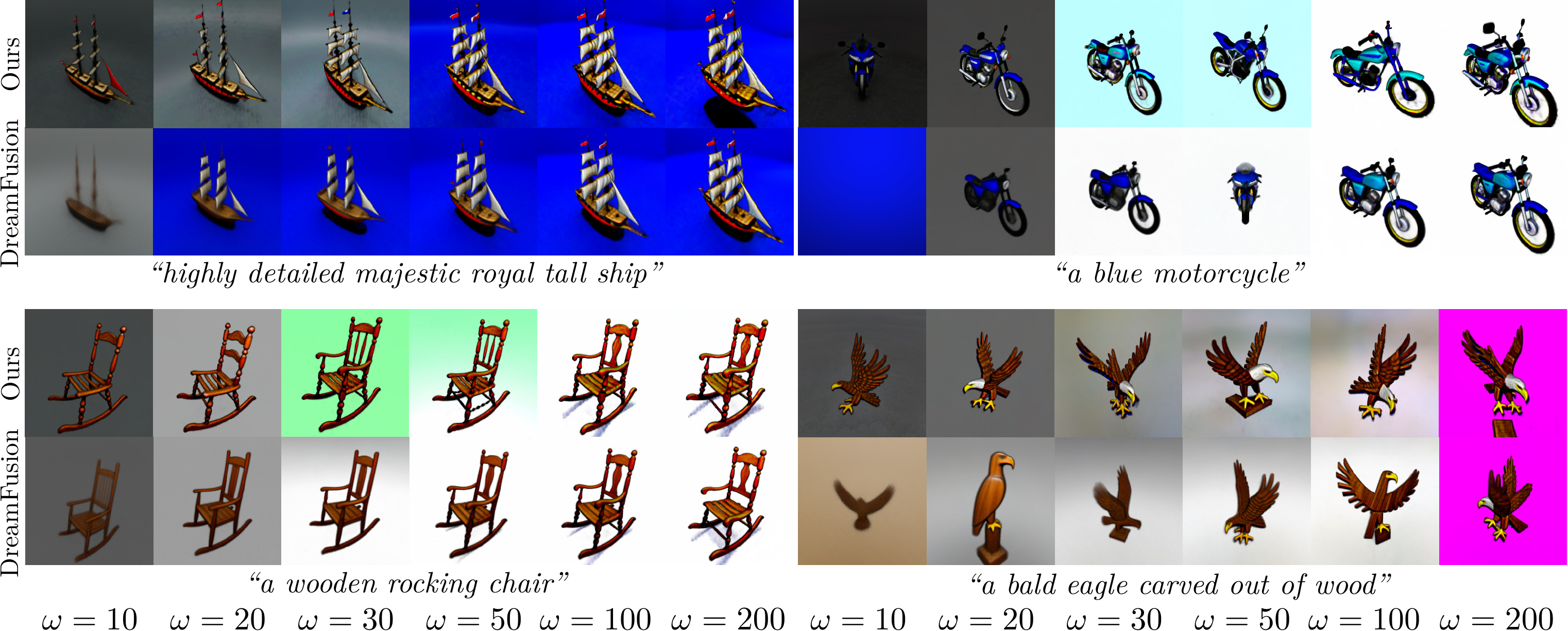}
    \caption{The effect of increasing guidance weight $\guidance$ in DreamFusion  using the original SDS loss (even rows) and our proposed LMC-SDS (odd rows). Using larger $\guidance$ saturates colors (chair, motorcycle, ship) and worsens the Janus problem (third wing and claw for the eagle). For low $\guidance$, vanilla DreamFusion fails to produce detailed geometry (ship, eagle) or geometry at all (motorcycle), while our approach performs well for $\guidance=20$, often even for $\guidance=10$. Notice how our approach constantly produces more detailed results for all $\guidance$.}
    \label{fig:guidance_ablation}
    \vspace{-4mm}
\end{figure}

\begin{figure}[t]

    \centering
    \includegraphics[width=\linewidth]{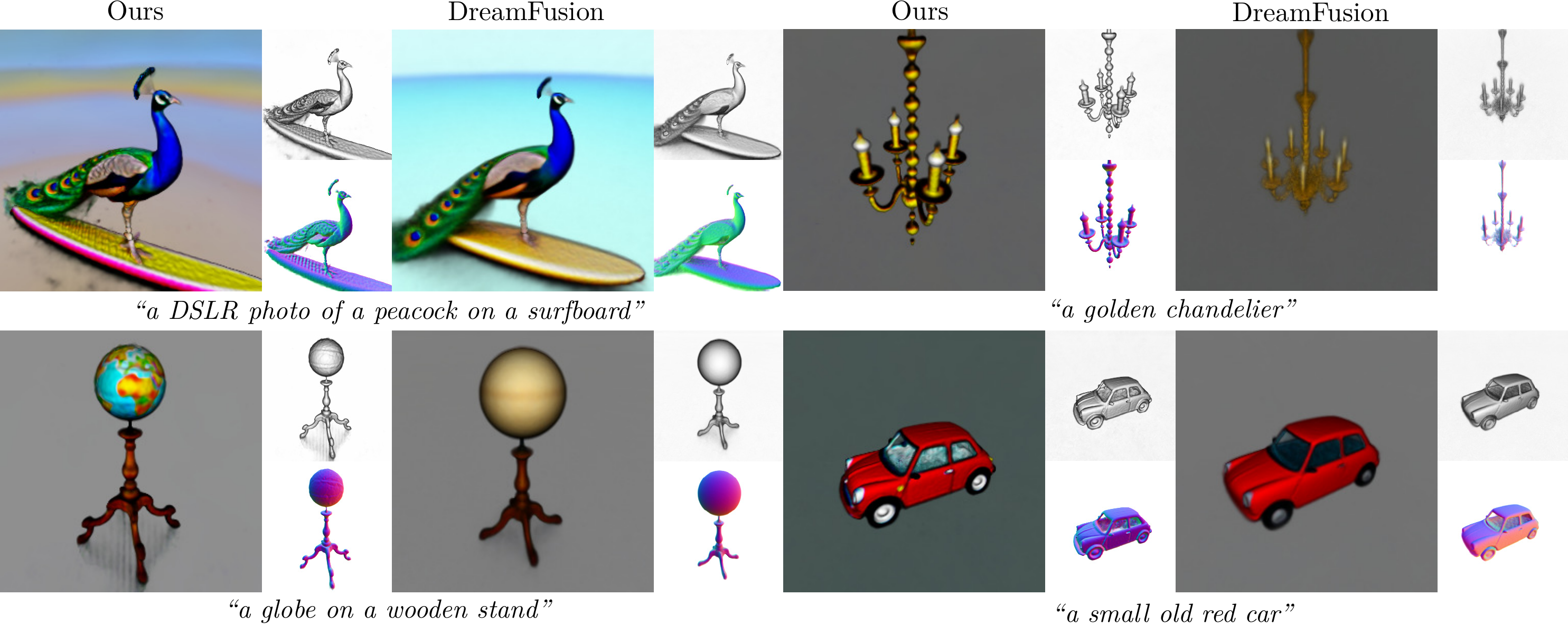}
    \caption{Results of DreamFusion using the original SDS loss (right) and our proposed LMC-SDS (left) for $\guidance=20$. Our results are much sharper, contain more detail, and feature more realistic colors in all examples.}
    \label{fig:dreamfusion_compare}
\end{figure}

\vspace{-2mm}
\subsection{Text-to-3D Asset Generation}
\vspace{-2mm}

\begin{figure}[t]
    \centering
    \includegraphics[width=\linewidth]{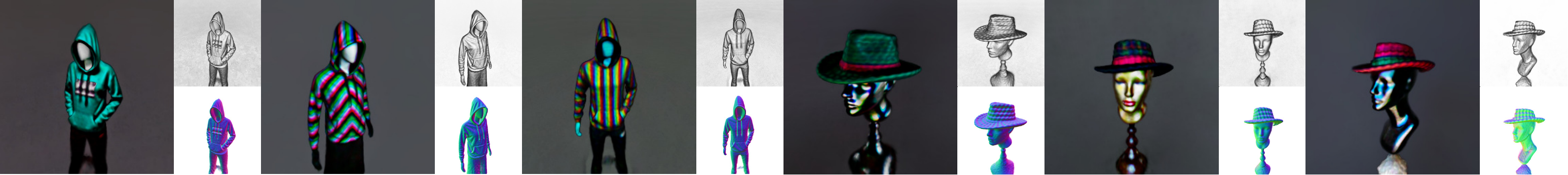}
    \caption{DreamFusion results using our LMC-SDS loss and fixed $\eps$ sampling for the prompts \emph{``a mannequin wearing a [hoodie, hat]''}. Note the diversity in colors, poses, and mannequin styles.}
    \label{fig:dreamfusion_seed}

\end{figure}

In our final experiment, we demonstrate how LMC-SDS can improve text-to-3D models in the style of DreamFusion.
In \cref{sec:analysis}, we discussed how large guidance weights $\guidance$ may affect the quality of results.
To provide further evidence for this observation, we produce 3D assets with varying $\guidance$ using the original DreamFusion formulation and a variant using our LMC-SDS.
The results in \cref{fig:guidance_ablation} also show the trend of high $\guidance$ having the tendency to produce oversaturated colors, albeit not as as strongly as in the original paper, possibly  due the use of a different image diffusion model.
But we  also observed an additional issue: the infamous Janus problem, the tendency to repeat detail in each view, becomes more severe with bigger $\guidance$, as evident from the eagle example.
On the other hand, when using low $\guidance$ in the original SDS loss, we observe that DreamFusion struggles to create detailed structure or even structure at all.
In contrast, when using LMC-SDS, DreamFusion successfully produces detailed results, even when using up to only one tenth of the originally proposed $\guidance$.
In \cref{fig:dreamfusion_compare} (more in Suppl.\ Mat.), we show additional side-by-side comparisons of the original SDS-based DreamFusion and our LMC-SDS version.
Using our loss, the results are consistently sharper and feature far more detail.
Finally in \cref{fig:dreamfusion_seed}, we show that fixed $\eps$ sampling may also be used in 3D asset generation, to obtain diverse results.

\vspace{-2mm}
\section{Discussion \& Conclusion}
\vspace{-2mm}
We proposed a novel loss formulation called LMC-SDS.
LMC-SDS allows to effectively utilize text-to-image diffusion models as rich image priors in several optimization and network training tasks.
We derived LMC-SDS by analyzing DreamFusion's SDS loss and by identifying an inherent problem in its formulation.
Concretely, we expressed the SDS gradient as the sum of two terms.
We identified that the second term $\loss_{\text{proj}}$ provides deficient gradients and consequently was weighted down in DreamFusion.
We argued that this term, however, should provide gradients towards the learned manifold of natural images associated to the diffusion model. We also provided evidence that the lack thereof makes the optimization unstable, thus negatively impacting the fidelity of results.
To this end, we proposed to model the denoising models' time step dependent frequency bias.
This allowed us to factor it out and obtain the cleaner gradient $\nabla_{\renderfnweights}\loss_{\text{lmc}}$.
In extensive experiments, we have demonstrated the effectiveness and versatility of our novel LMC-SDS loss.
Furthermore, we discussed how LMC-SDS differs from other SDS-style losses and demonstrated how its performance is superior both qualitatively and quantitatively.

\paragraph{Limitations.}
LMC-SDS finds its limits where either the signal from the diffusion model is too weak and ambiguous or when the optimization goes off the natural image manifold by too much.
Concretely, LMC-SDS can only utilize whatever signal is coming from the diffusion model.
This means, that when the diffusion model does not ``understand'' the prompt, LMC-SDS (and all other SDS-based approaches) will not be able to provide meaningful gradients.
Further, whenever the current optimization state is too corrupted, LMC-SDS will not be able to correct for that and guide the optimization back to the manifold.
Also, LMC-SDS sometimes overcompensates the frequency bias. 
Nevertheless, LMC-SDS is an important first step towards more stable and meaningful usage of the inherent image prior associated to image diffusion models.

We proposed LMC-SDS, a novel SDS-based loss formulation with cleaner gradients towards the image manifold.
In the future, we would like to continue to investigate how the manifold corrective can be further improved. We also want to leverage our findings in concrete applications like text-to-3D or  in image editing and stylization tasks.

\appendix
\section*{\Large Supplementary Material}

In this supplementary material, we detail the  training of $\approxbias_{\kernelweights}$ and the implementation and hyper-parameters of our experiments.
We also present additional results.

\vspace{-2mm}
\section{Implementation Details}
\vspace{-2mm}

In all experiments, we set $w(\timestep) = 1$, we sample $\timestep \sim \mathcal{U}(0.02, 0.98)$, and use the Adam optimizer. The only exceptions are the DreamFusion experiments, where we use the original hyper-parameters.

\paragraph{Training $\approxbias_{\kernelweights}$.}
We model $\approxbias_{\kernelweights}$ as a standard U-Net with nine layers with two Conv/ReLU/MaxPool blocks per layer, and skip connections from the encoder to the decoder.
The first layer contains 32 filters and the number of filters is doubled in each layer in the encoder and halved in the decoder.
The output is computed with a final Conv layer with $\tanh$ activation.
The time step $\timestep$ is passed via concatenation to each pixel and encoded using 10-dimensional frequency encoding.
$\approxbias_{\kernelweights}$ has been trained until convergence at $\sim750$K steps, using $\loss_{\text{k}}$ as the only loss, a batch size of $128$, and with a learning rate of $1\times10^{-4}$ decayed by $0.9$ every 100K steps.

\paragraph{Image synthesis.}
We initialize an image of $512 \times 512$px resolution with 50\% grey. We then optimize for 500 iterations using 
$\loss_{\text{LMC-SDS}} = \loss_{\text{lmc}} + \loss_{\text{cond}}$ with $\guidance=8$ and using a learning rate of $0.1$ with Cosine decay of $0.03$ over 300 steps. We scale the whole image to $128\times128$px for the first 150 steps and then switch to our patching strategy, where we sample two patches in each iteration.

\paragraph{Image editing.}
We initialize an optimizable image of $512 \times 512$px resolution with pixel values from the reference image. We then optimize for 500 iterations using 
$\loss_{\text{LMC-SDS}}$ with $\guidance=15$ and an L2 loss on the original image weighted by $0.1$.
We use a learning rate of $0.02$ with Cosine decay of $0.8$ over all steps and optimize the whole image for 200 steps before switching to patches.
The numerical experiment has been conducted on $256 \times 256$px images. Also, we didn't use L2 regularization and thus lowered the initial learning rate to $5\times10^{-3}$.

\paragraph{Image-to-image translation network.}
We trained a standard U-Net with 11 layers with two Conv/ReLU/MaxPool blocks per layer, and skip connections from the encoder to the decoder. The first layer contains 32 filters and the number of filters is doubled until 256 in each layer in the encoder and halved in last layers of the decoder. 
The class id is a float value encoded using 10-dimensional frequency encoding concatenated to each pixel.
The final Conv layer with $\tanh$ activation outputs images of $256 \times 256$px resolution.
We trained the network with a learning rate of $5\times 10^{-5}$ with a decay of $0.9$
every $50$K steps.
In the first $20$K steps, we use Cosine scheduling to increase the weight of $\loss_{\text{LMC-SDS}}$ from $0.1$ to $1.0$ and decrease the weight of L2 regularization \wrt the original image from $1.0$ to $0.1$.
We trained for $200$K iterations with batches of $32$ images.

\paragraph{Text-to-3D asset generation.} We used the original DreamFusion implementation and kept all hyper-parameters.

\paragraph{Metrics.} We report CLIP scores based on the CLIP ViT-B-16 model. LPIPS metrics are computed with the original AlexNet variant.

\vspace{-2mm}
\section{Additional Results}
\vspace{-2mm}

\begin{figure}[t]
    \centering
    \includegraphics[width=\linewidth]{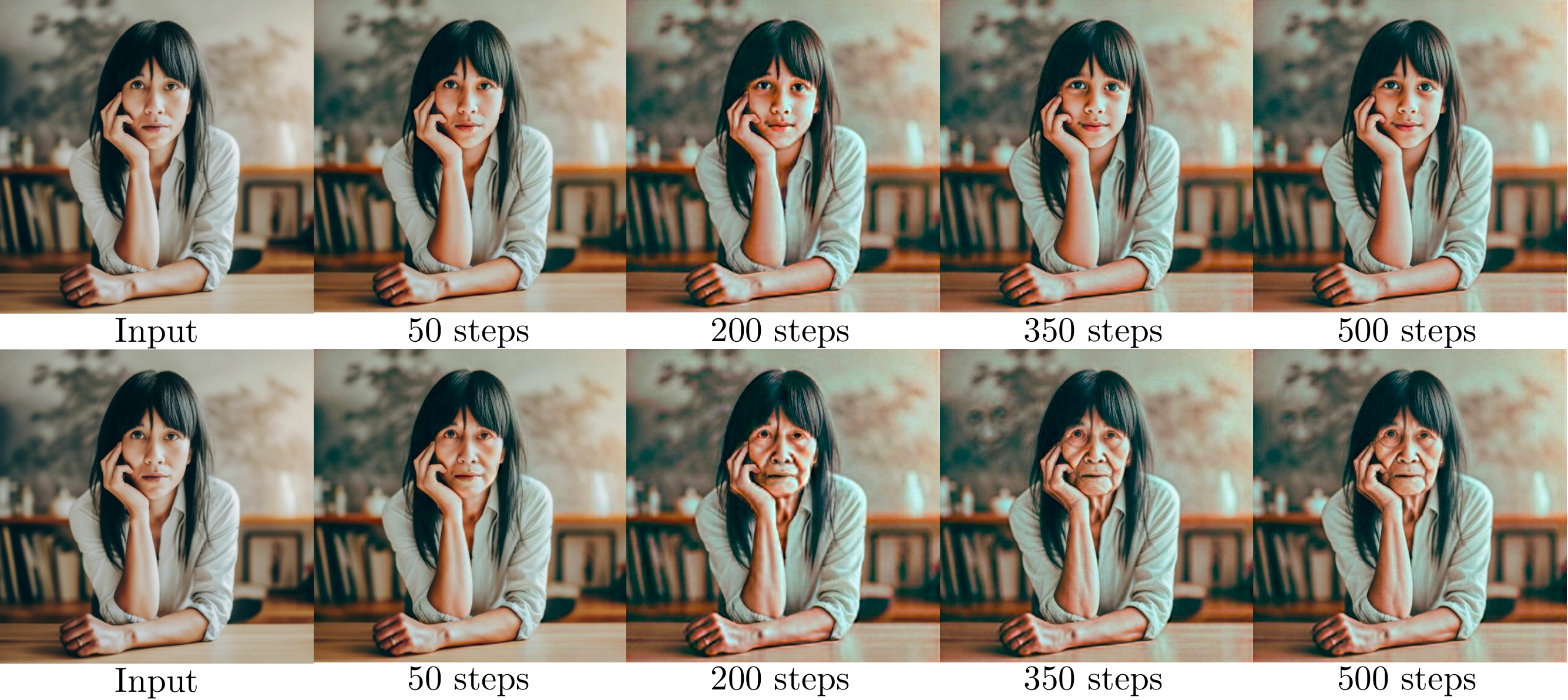}
    \caption{Progression of an image optimized using the prompts \emph{``photo of a young girl''} (top) and \emph{``photo of an old woman''} (bottom).}
    \label{fig:suppl_progress}
\end{figure}

\begin{figure}[t]

    \centering
    \includegraphics[width=\linewidth]{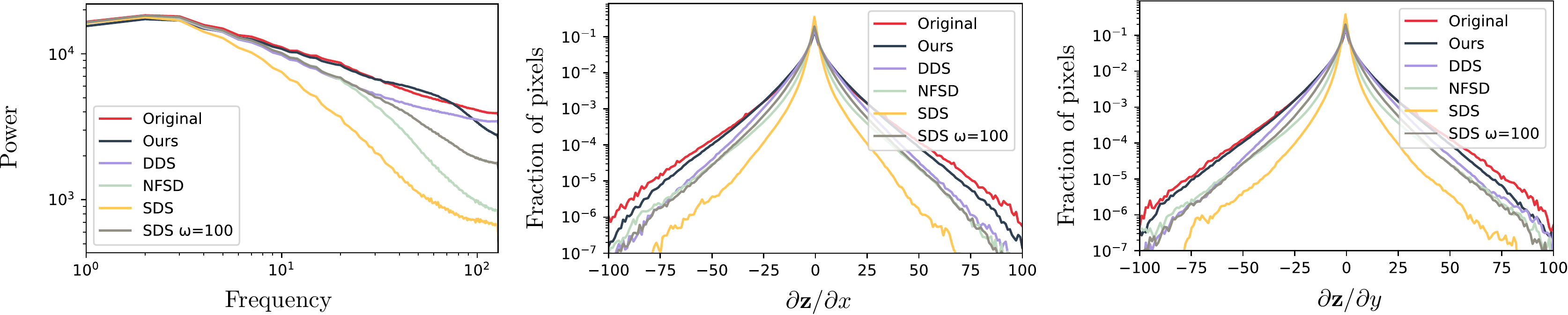}
    \caption{Image statistics of the results of the various methods in the \emph{cats-to-others} experiment. The statistics correlate well with the numerical and qualitative results reported in the main paper. SDS introduces significant blur to the results, while our method preserves the original statistics best.}
    \label{fig:suppl_statistics}
\vspace{8mm}
    \centering
    \includegraphics[width=\linewidth]{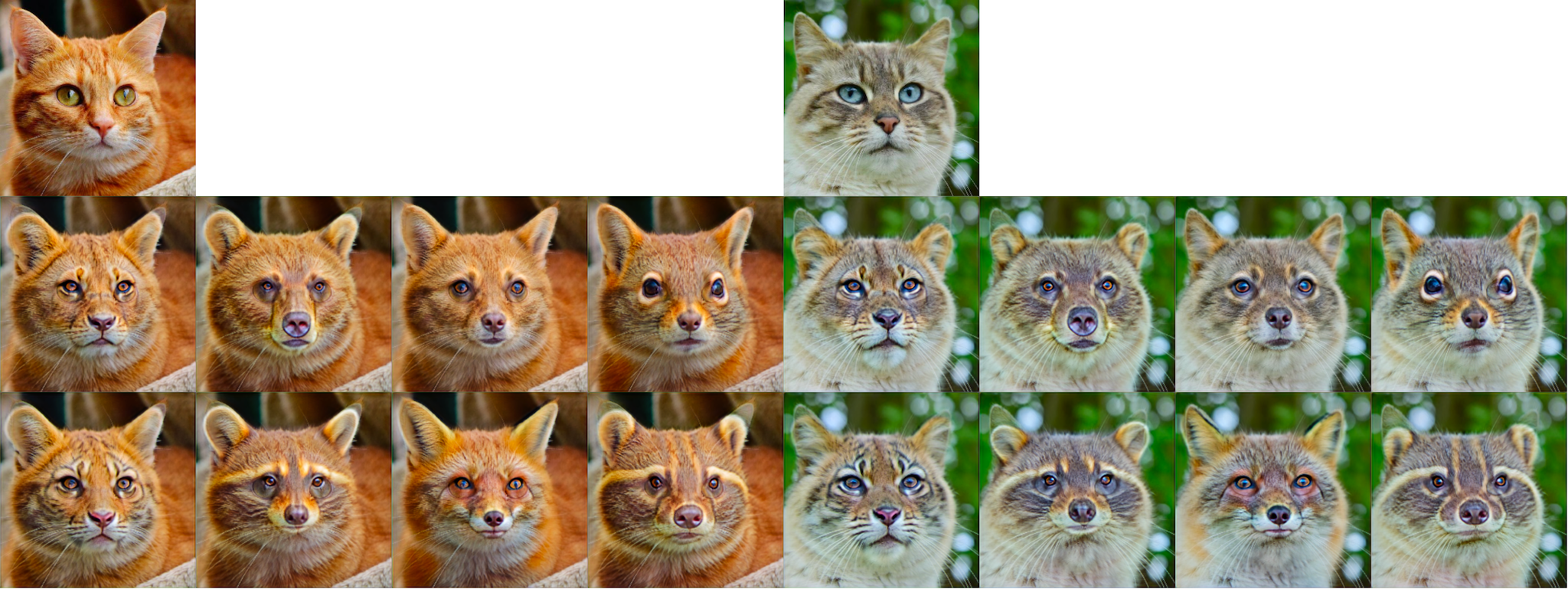}
    \caption{Qualitative results of a \emph{cats-to-others} image translation
network trained on eight classes. First row: Input image. Middle row: Results for \emph{``lion''}, \emph{``bear''}, \emph{``dog''}, and \emph{``squirrel''}. Last row: Results for \emph{``tiger''}, \emph{``raccoon''}, \emph{``fox''}, and \emph{``badger''}.}
    \label{fig:suppl_8_classes}
\vspace{8mm}
    \centering
    \includegraphics[width=\linewidth]{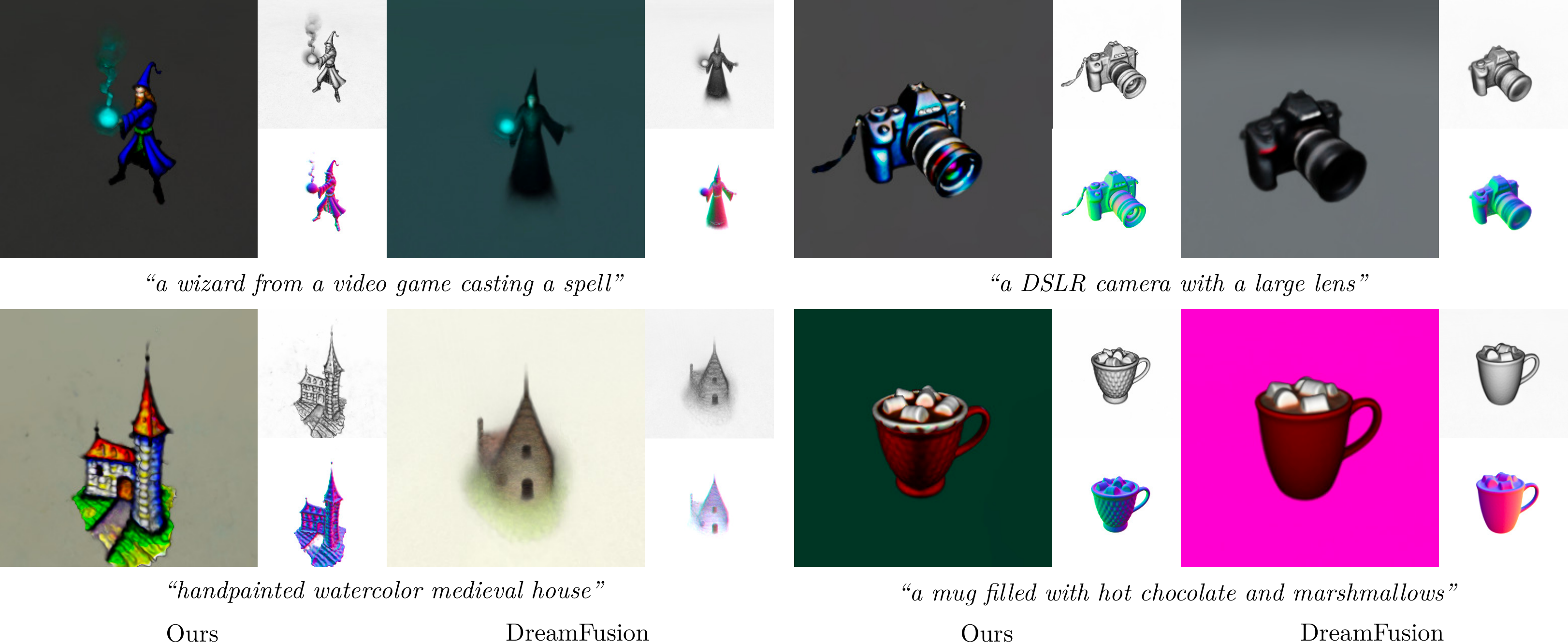}
    \caption{Additonal results of DreamFusion using the original loss formulation and our proposed LMC-SDS (Ours) for $\guidance=20$. Results obtained using our LMC-SDS are much sharper and contain more detail in all examples.}
    \label{fig:suppl_dreamfusion}
\end{figure}

In \cref{fig:suppl_progress}, we visualize the progression of optimization-based image editing. Note how intermediate stages are all realistic natural images.
Our LMC loss component provides a gradient towards the natural image manifold.
Thus, during optimization, the image always stays close to that manifold.

\begin{figure*}
    \centering
    \scriptsize
    \setlength{\tabcolsep}{0pt} %
    \renewcommand{\arraystretch}{0.0}
    \newcolumntype{Y}{>{\centering\arraybackslash}X}
    \begin{tabularx}{0.7\textwidth}{@{}YYYY@{}}
    \multicolumn{4}{c}{\includegraphics[width=0.7\linewidth]{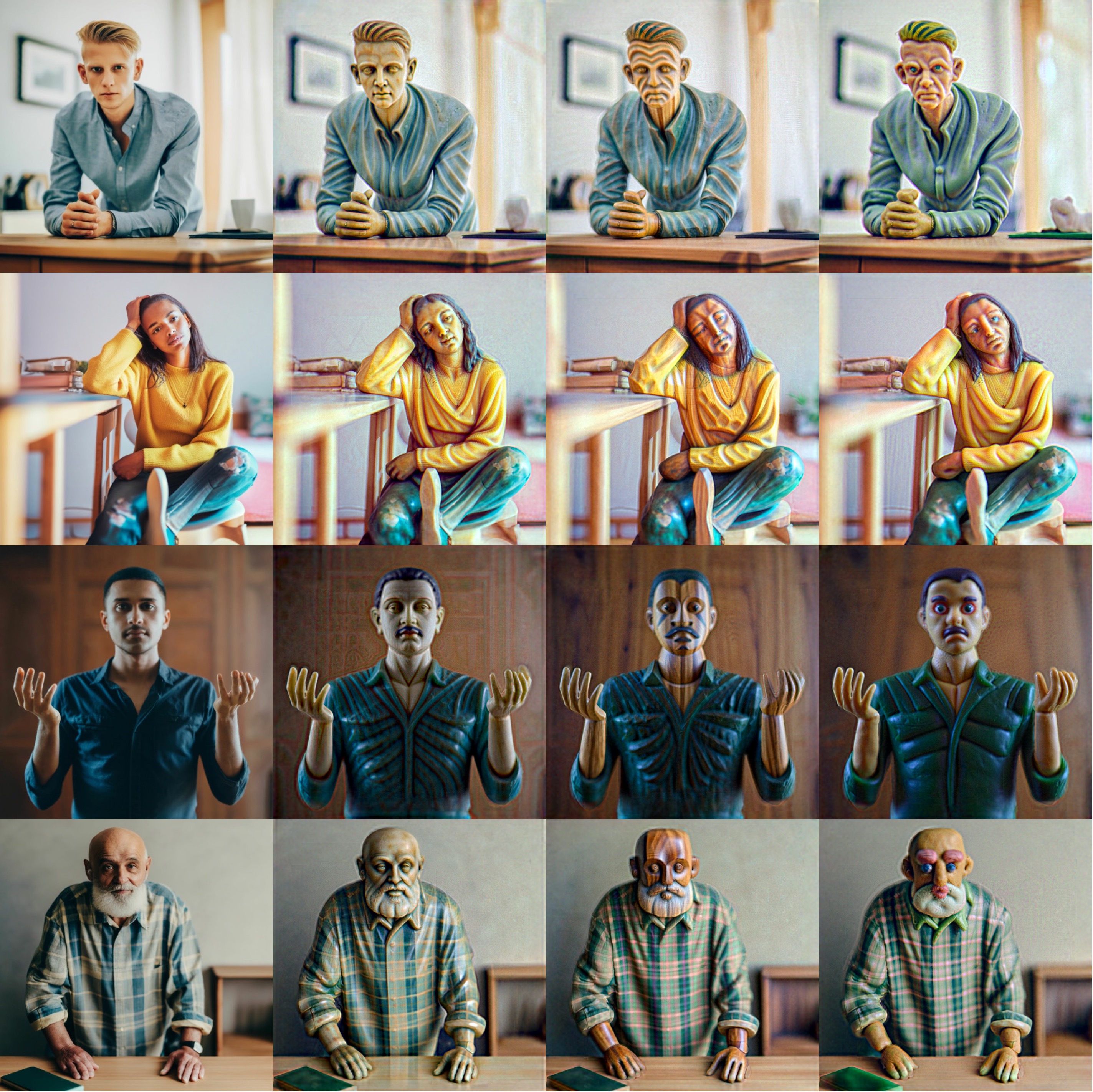}}\\%
    \multicolumn{4}{c}{\includegraphics[width=0.7\linewidth]{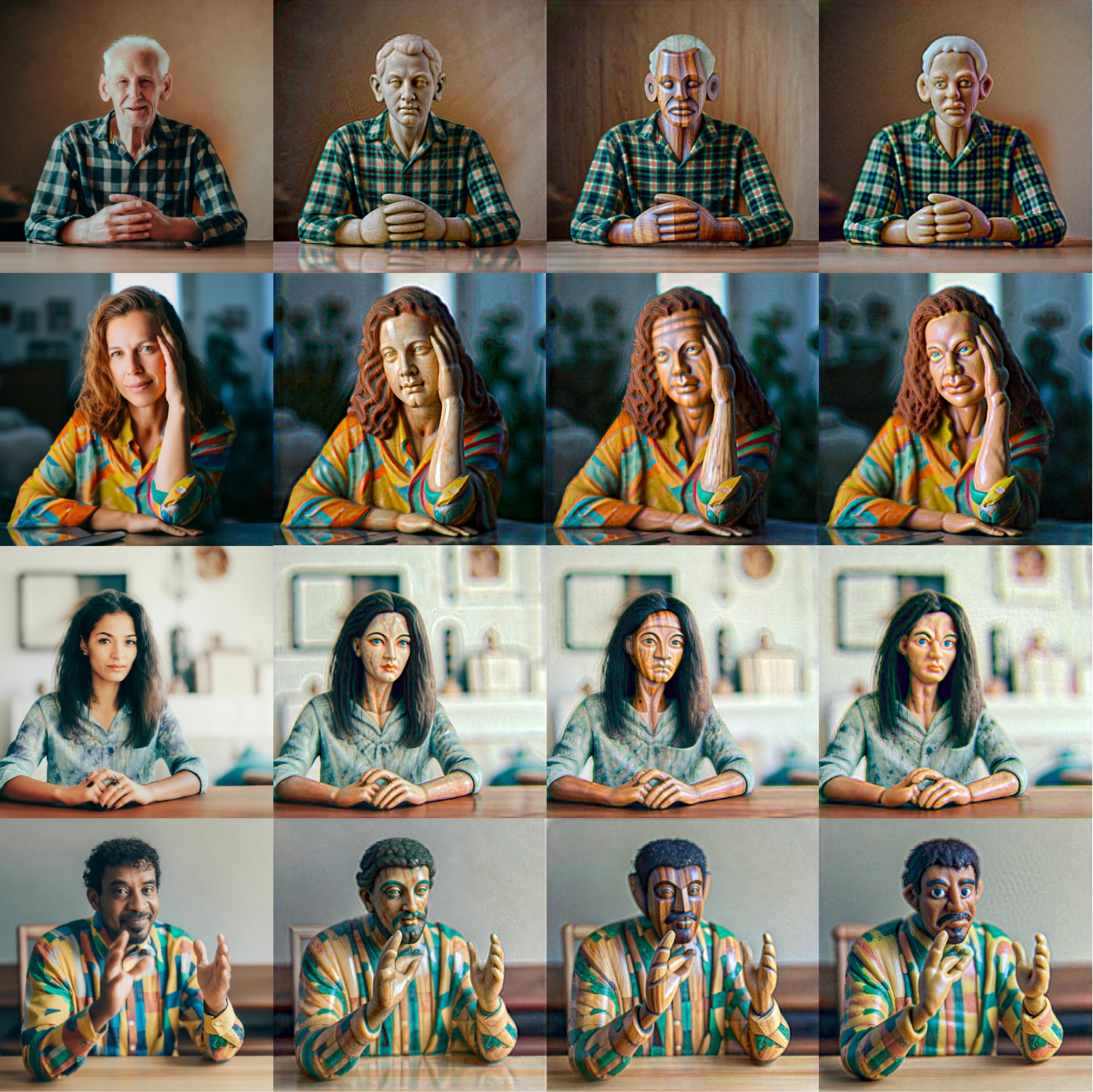}}\vspace{3pt}\\
        Input & \emph{``marble''} & \emph{``wood''} & \emph{``clay''} \\
    \end{tabularx}
    \caption{Optimization-based editing results for prompts describing different materials. The full prompts are \emph{``a marble statue, made from marble, stone''}, \emph{``a wood sculpture, wooden statue, made from wood''}, \emph{``clay figure, made from modelling clay''}.}
    \label{fig:suppl_art}
\end{figure*}

\begin{figure*}
    \centering
    \scriptsize
    \setlength{\tabcolsep}{0pt} %
    \renewcommand{\arraystretch}{0.0}
    \newcolumntype{Y}{>{\centering\arraybackslash}X}
    \begin{tabularx}{0.7\textwidth}{@{}YYYY@{}}
    \multicolumn{4}{c}{\includegraphics[width=0.7\linewidth]{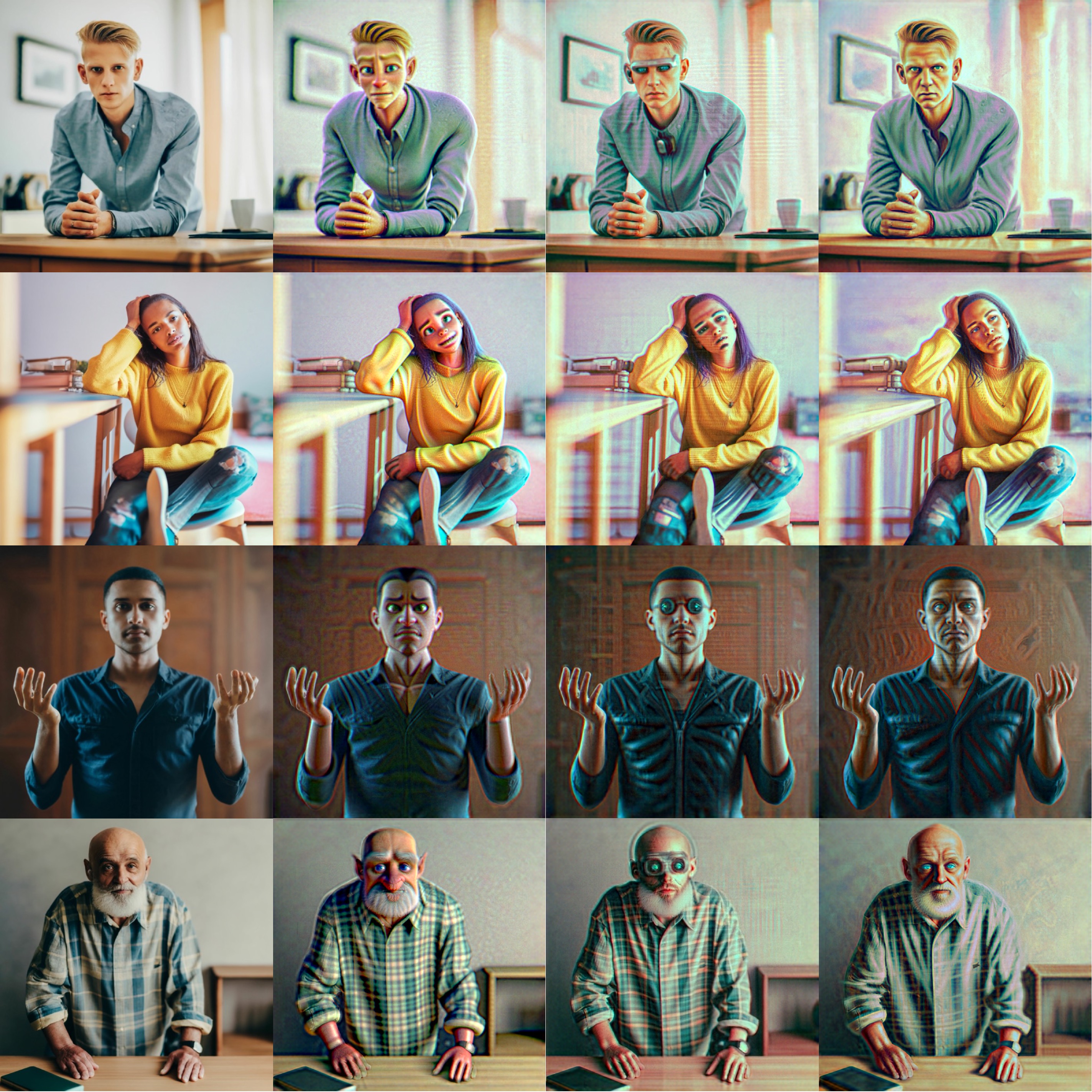}}\\%
    \multicolumn{4}{c}{\includegraphics[width=0.7\linewidth]{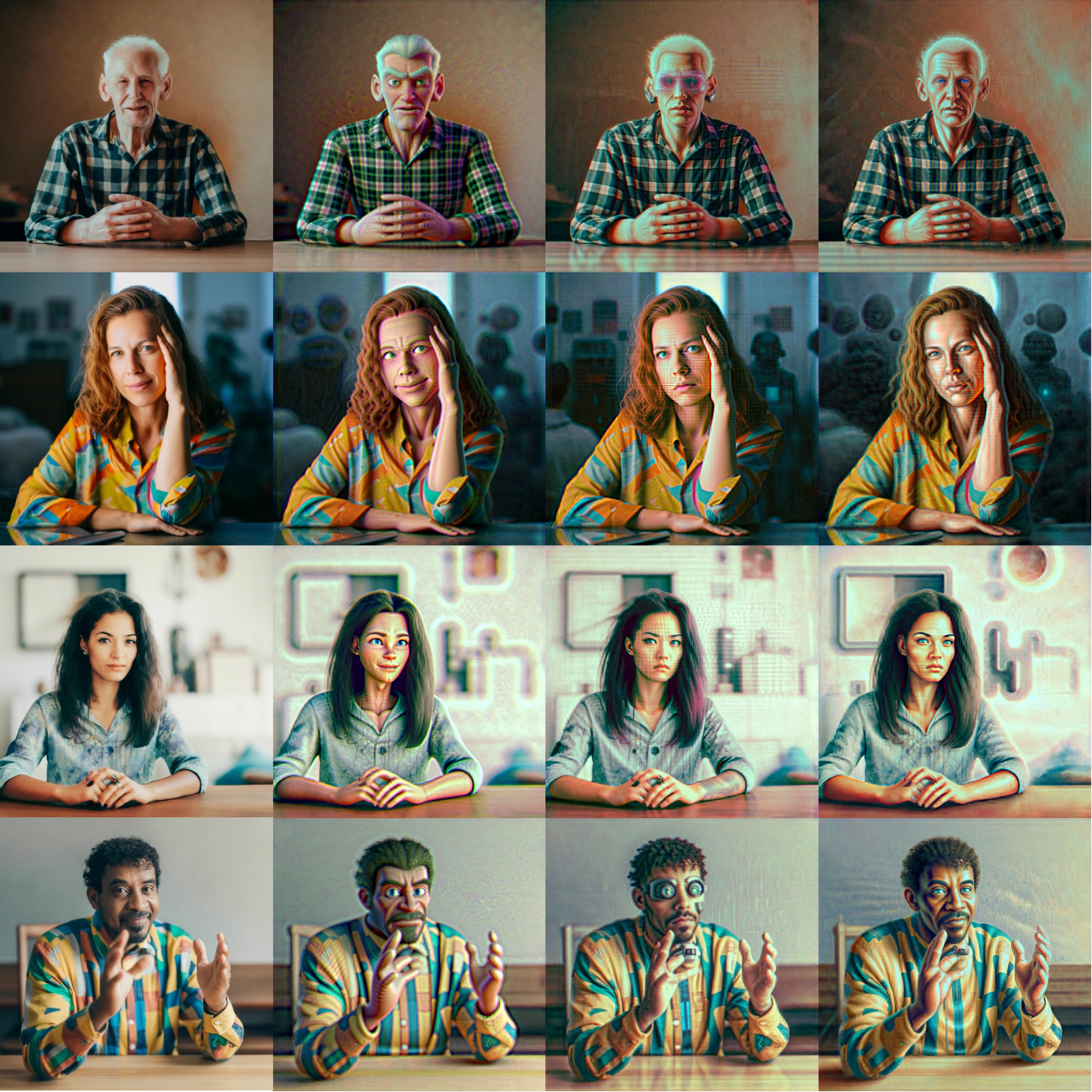}}\vspace{3pt}\\
        Input & \emph{``3D animation''} & \emph{``cyberpunk''} & \emph{``digital painting''} \\
    \end{tabularx}
    \caption{Optimization-based editing results for prompts describing different digital art genres. The full prompts are \emph{``a character from a 3D animation movie''}, \emph{``cyberpunk, futuristic, dystopian''}, and \emph{``sci-fi, digital painting''}.}
    \label{fig:suppl_art_2}
\end{figure*}

\begin{figure*}
    \centering
    \scriptsize
    \setlength{\tabcolsep}{0pt} %
    \renewcommand{\arraystretch}{0.0}
    \newcolumntype{Y}{>{\centering\arraybackslash}X}
    \begin{tabularx}{0.7\textwidth}{@{}YYYY@{}}
    \multicolumn{4}{c}{\includegraphics[width=0.7\linewidth]{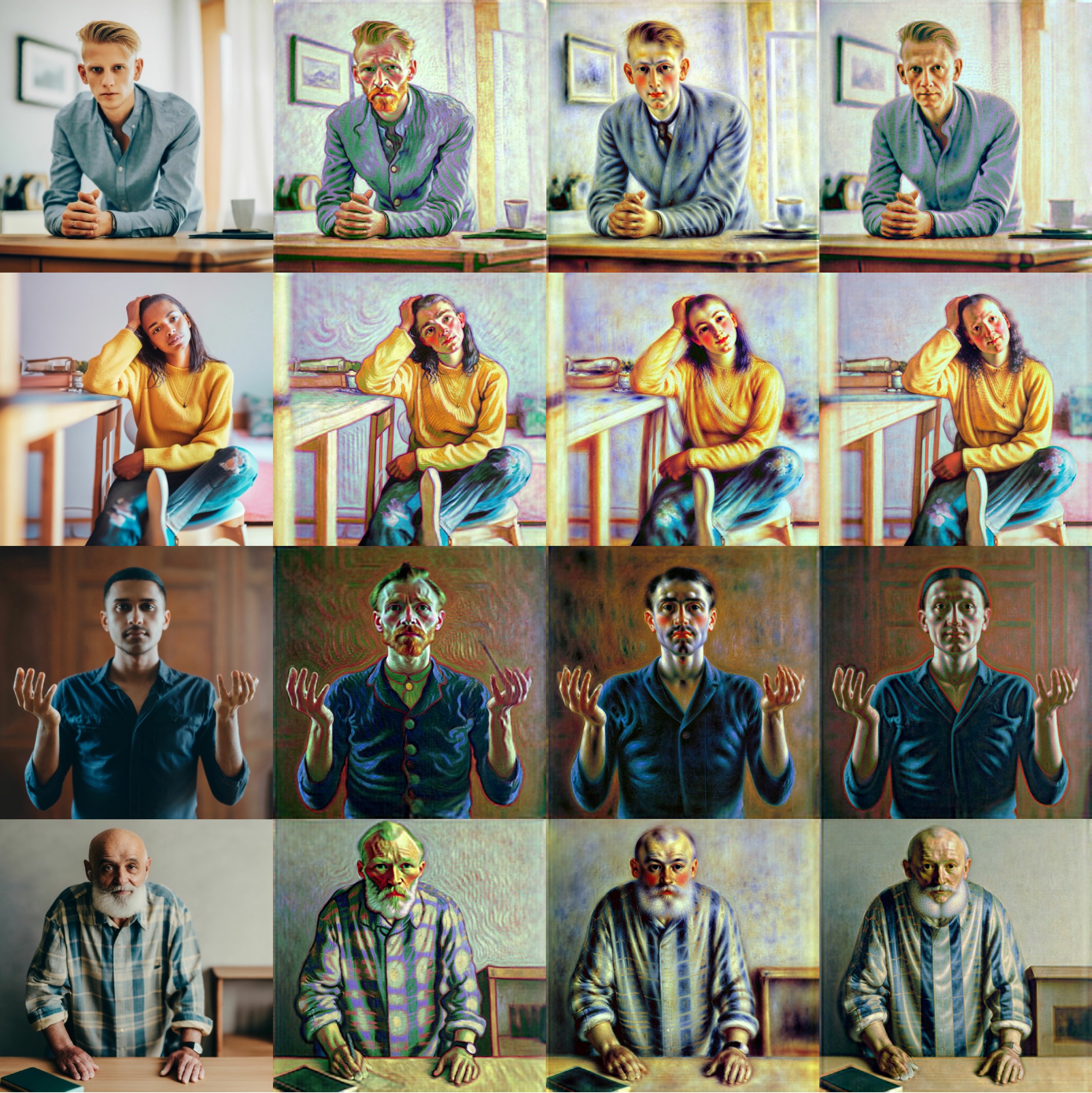}}\\%
    \multicolumn{4}{c}{\includegraphics[width=0.7\linewidth]{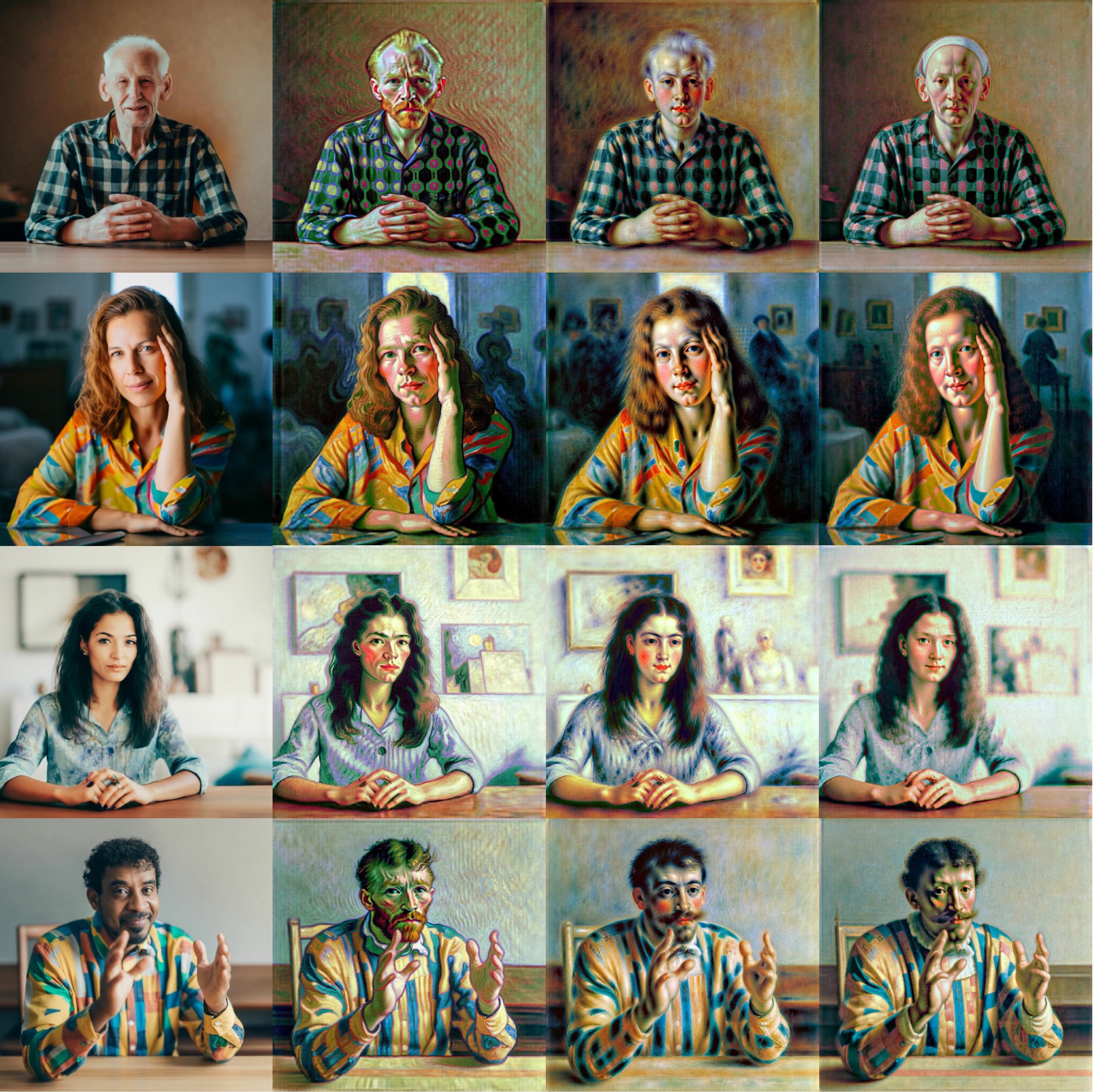}}\vspace{3pt}\\
        Input & \emph{``van Gogh''} & \emph{``Renoir''} & \emph{``Vermeer''} \\
    \end{tabularx}
    \caption{Optimization-based editing results for prompts describing different artists. The full prompts are \emph{``art by Vincent van Gogh''}, \emph{``art by Pierre-Auguste Renoir''}, \emph{``art by Johannes Vermeer''}.}
    \label{fig:suppl_art_3}
\end{figure*}

\begin{figure*}
    \centering
    \scriptsize
    \setlength{\tabcolsep}{0pt} %
    \renewcommand{\arraystretch}{0.0}
    \newcolumntype{Y}{>{\centering\arraybackslash}X}
    \begin{tabularx}{0.7\textwidth}{@{}YYYY@{}}
    \multicolumn{4}{c}{\includegraphics[width=0.7\linewidth]{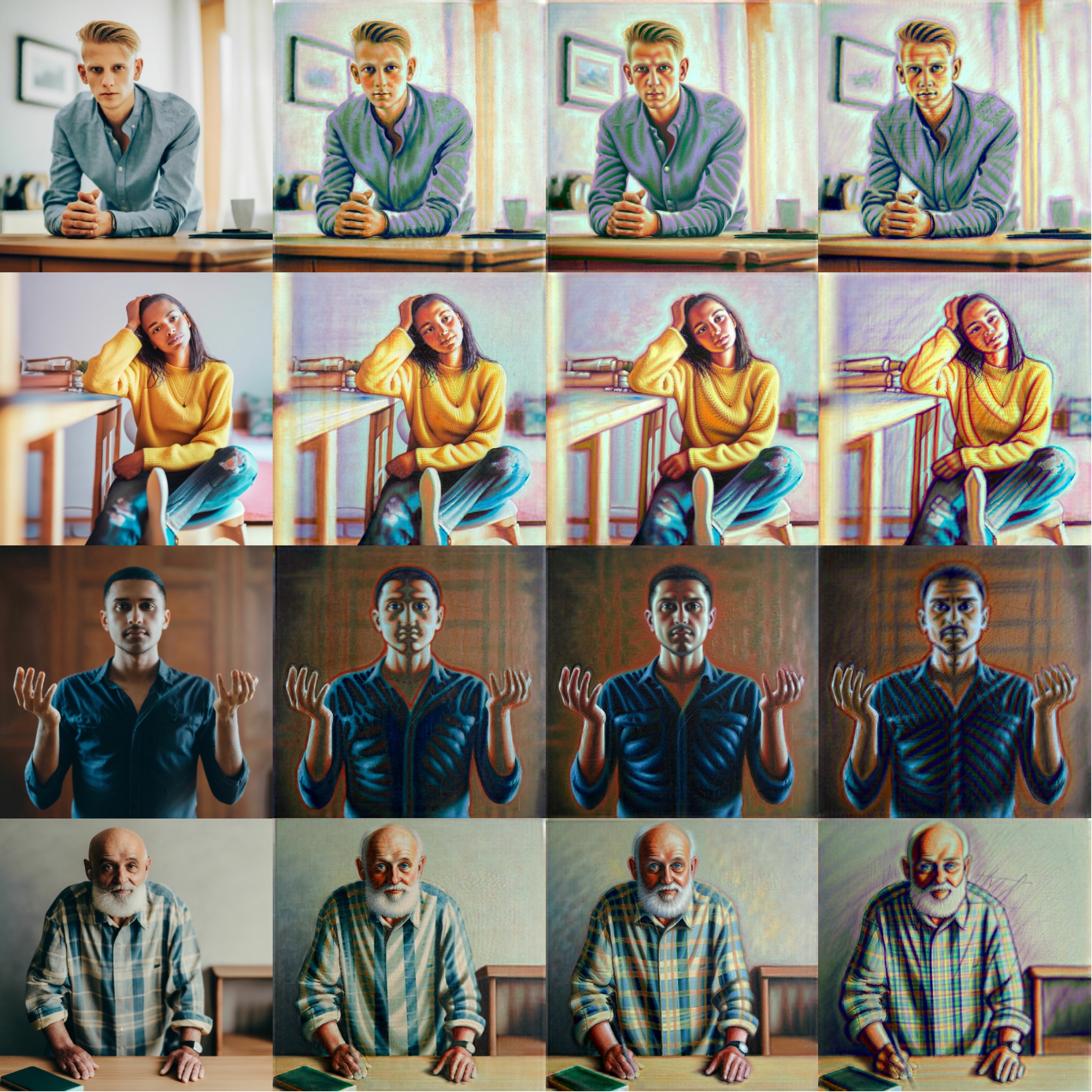}}\\%
    \multicolumn{4}{c}{\includegraphics[width=0.7\linewidth]{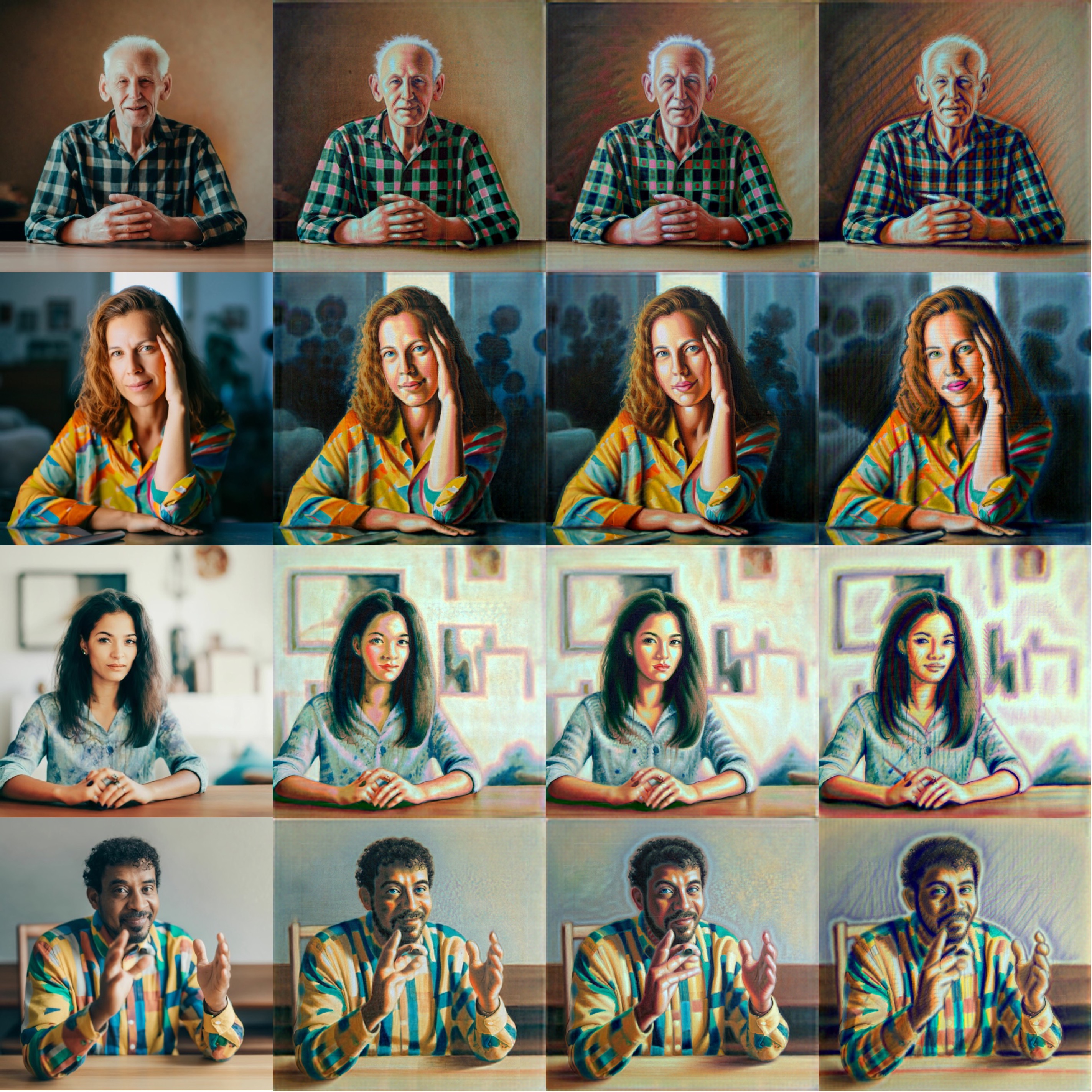}}\vspace{3pt}\\
        Input & \emph{``aquarelle''} & \emph{``oil''} & \emph{``pencil''} \\
    \end{tabularx}
    \caption{Optimization-based editing results for prompts describing different art techniques. The full prompts are \emph{``aquarelle painting''}, \emph{``oil painting''}, and \emph{``pencil sketch''}.}
    \label{fig:suppl_art_4}
\end{figure*}

In \cref{fig:suppl_art,fig:suppl_art_2,fig:suppl_art_3,fig:suppl_art_4} we present additional editing results. 
In contrast to the main paper, we show results for each prompt on a number of images. Concretely, we synthesized photographs of people and edited each of them using prompts describing different art styles or materials.
Each result respects the corresponding prompt well and yet has an individual touch to it.
The image statistics are well preserved and artifacts are rare. 
The overall quality once more demonstrates the effectiveness of our novel LMC-SDS loss.

We report image statistics of the results of all methods in the \emph{cats-to-others} network training experiment in \cref{fig:suppl_statistics}.
In \cref{fig:suppl_8_classes}, we repeated the experiment with our LMC-SDS, but used eight classes instead of four. 
We did not increase the network capacity or fine-tuned hyper-parameters.
Nevertheless, the trained network is able to perform convincing edits on all eight classes.

Finally, in \cref{fig:suppl_dreamfusion} we show additional comparisons of DreamFusion results using our novel LMC-SDS and the vanilla SDS loss, both using $\guidance=20$.

\newpage
\onecolumn
\bibliographystyle{splncs04}

\end{document}